%% file: iclr2024_conference.tex
\documentclass{article} % For LaTeX2e
\usepackage{iclr2024_conference,times}

\input{math_commands.tex}

\usepackage{hyperref}
\usepackage{url}

\usepackage[inline,shortlabels]{enumitem}
\usepackage{amsmath}
\usepackage{amssymb}
\usepackage{mathtools}
\usepackage{amsthm}
\usepackage{comment}
\usepackage{xspace}
\usepackage{multirow}
\usepackage{wrapfig}
\usepackage[font=small,labelfont=bf]{caption}
\usepackage{graphicx}
\usepackage{subfigure}
\usepackage[ruled, lined, linesnumbered, commentsnumbered, longend]{algorithm2e}
\usepackage{booktabs}       % professional-quality tables
\newcommand{\SpaSB}{\textsc{BrusLeAttack}\xspace} %BayPix
\newcommand{\SparseRS}{\textsc{Sparse-RS}\xspace}
\newcommand{\SpaEvoAtt }{\textsc{SparseEvo}\xspace}
\newcommand{\pgdlO}{$\text{PGD}_0$\xspace}
\newcommand{\bo}{\textsc{CASMOPOLITAN}\xspace}
\newcommand{\hl}[1]{\textcolor{black}{#1}}

\def\ie{\emph{i.e.}\xspace}

\title{\textsc{BruSLeAttack}: A Query-Efficient Score-Based Black-Box Sparse Adversarial Attack}

\author{Viet Quoc Vo, Ehsan Abbasnejad, Damith C. Ranasinghe\\
The University of Adelaide\\
\texttt{\{viet.vo,ehsan.abbasnejad,damith.ranasinghe\}@adelaide.edu.au}}

\iclrfinalcopy % Uncomment for camera-ready version, but NOT for submission.
\begin{document}

\maketitle
\begin{abstract}
 We study the unique, less-well understood problem of generating \textit{sparse adversarial} samples \textit{simply} by observing the \textit{score-based} replies to \textit{model queries}. Sparse attacks aim to discover a \textit{minimum} number---the $l_0$ bounded---perturbations to model inputs to craft adversarial examples and \textit{misguide} model decisions. But, in contrast to query-based dense attack counterparts against black-box models, constructing sparse adversarial perturbations, even when models serve \textit{confidence score information} to queries in a \textit{score-based} setting, is non-trivial. Because, such an attack leads to: i)~an NP-hard problem; and ii)~a non-differentiable search space. We develop the \textsc{\SpaSB}---a \textit{new}, \textit{faster} (more query efficient) Bayesian algorithm for the problem. We conduct extensive attack evaluations 
 including an \textit{attack demonstration} against a Machine Learning as a Service (MLaaS) offering exemplified by {\small{\textsf{\textbf{Google Cloud Vision}}}} and robustness testing of adversarial training regimes and a recent defense against black-box attacks. The proposed attack scales to achieve \textit{state-of-the-art attack success rates} and \textit{query efficiency} on standard computer vision tasks such as {\small{\texttt{\textbf{ImageNet}}}} across different model architectures. 
 Our artifacts and DIY attack samples are available on \href{https://BruSLiAttack.github.io}{GitHub}. Importantly, our work facilitates \textit{faster} evaluation of model vulnerabilities and raises our vigilance on the safety, security and reliability of deployed systems.
 \end{abstract}

%%%%%%%%% BODY TEXT
\section{Introduction}
\label{sec:intro}

%-------------------------------------------------------------------------------
We are amidst an increasing prevalence of deep neural networks in real-world systems. So, our ability to understand the safety and security of neural networks is critical to our \textit{trust} in machine intelligence. We have heightened awareness of adversarial attacks~\citep{Szegedy2013}---crafting imperceptible perturbations in inputs to manipulate deep perception systems to produce erroneous decisions. In real-world applications such as %autonomous cars and 
machine learning as a service (MLaaS) from {\small\textsf{Google Cloud Vision}} or {\small\textsf{Amazon Rekognition}}, the model is \textit{hidden} from users. Only, access to model decisions (labels) or confidence \textit{scores} are possible. Thus, crafting adversarial examples in black-box \textit{query-based} interactions with a model is both interesting and practical to consider.

\begin{figure}[htb]
    \begin{center}
        \includegraphics[scale=0.70]{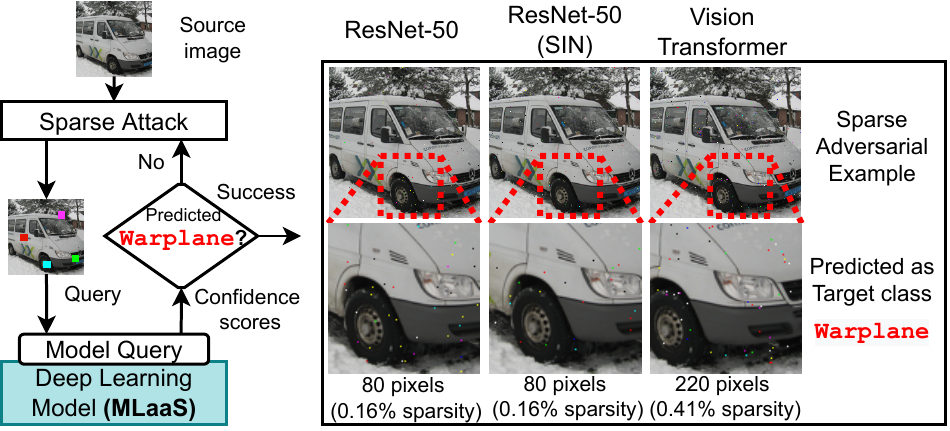}
        \caption{\textbf{\textit{Targeted Attack}}. Malicious instances are generated by \SpaSB with different perturbation budgets against three Deep Learning models on \texttt{ImageNet}. An image with ground-truth label \texttt{Minibus} is misclassified as a \texttt{Warplane}. Interestingly, in contrast to needing \textit{220 pixels} to mislead the Vision Transformer, \SpaSB requires only \textit{80 perturbed pixels} to fool ResNet-based models (more visuals in \textbf{Appendix}~\ref{apdx:Visualization of Sparse Adversarial Examples}). %Visualizations and 
        Evaluation against {\small\textsf{\textbf{Google Cloud Vision}}} is in \textbf{Section}~\ref{subsec:Attack a Real-World System} and \textbf{Appendix} \ref{apdx:Visualizations of Attack Against Google Cloud Vision}.}
        \label{fig:illustration of attack different DL models}
    \end{center}
    \vspace{-5mm}
\end{figure}

\textbf{Why Study \textit{Query}-Based \textit{Sparse} Attacks Under \textit{Score}-Based Responses?~}
Since confidence scores expose more information compared to model decisions, we can expect fewer queries to elicit effective attacks and, consequently, the potential for developing \textit{attacks at scale} under \textit{score-based} settings. 
%===================================
Various similarity measures---$l_p$ norms---are used to quantitatively describe adversarial example perturbations. Particularly, $l_2$ and $l_\infty$ norm is used to quantify \textbf{\textit{dense}} perturbations for attacks. In contrast, $l_0$ norm quantifies \textit{sparse perturbations} aiming to perturb a \textbf{\textit{tiny}} portion of the input. While dense attacks are widely explored, the success of \textit{sparse-attacks}, especially under \textit{score-based} settings, has drawn much less attention and remains less understood~\citep{Croce2022}. This leads to our lack of knowledge of model vulnerabilities to sparse perturbation regimes. 

\textbf{Why are \textit{Score}-Based \textit{Sparse} Attacks Hard?~}Constructing sparse perturbations is incredibly difficult as minimizing $l_0$ norm leads to an NP-hard problem~\citep{Modas2019, Dong2020} and a non-differentiable search space that is mixed (discrete and continuous)~\citep{Carlini2017}. Now, for a given $l_0$ constraint or number of pixels, we need to search for both the optimal set of pixels to perturb in a source image and the pixel colors—-floats in [0, 1]. Solutions are harder, if we aim to achieve both \textit{query efficiency} and \textit{high attack success rate} (ASR) for high resolution vision tasks such as \texttt{ImageNet}. The only \textit{scalable} attempt to the challenges, \SparseRS~\citep{Croce2022}, applies a stochastic search method to seek potential solutions.

\textbf{Our Proposed Algorithm.~}We consider a \textit{new formulation} to cope with the problem and construct the new search method--\SpaSB. We propose a search for a sparse adversarial example over an effective, lower dimensional search-space. In contrast to the prior stochastic search and pixel selection method, we guide the search by prior knowledge learned from historical information of pixel manipulations (past experience) and informed selection of pixel level perturbations from our lower dimensional search space to tackle the resulting combinatorial optimization problem. 

\textbf{Contributions.~}Our efforts increase our understanding of \textit{less-well} understood, \textit{hard}, score-based, query attacks to generate \textit{sparse} adversarial examples. Notably, only a few studies exist on the robustness of vision Transformer (ViT) architectures to sparse perturbation regimes. This raises a critical concern over their reliable deployment in applications. Therefore, we investigate the fragility of both CNNs and ViTs against sparse adversarial attacks. Figure \ref{fig:illustration of attack different DL models} demonstrates examples of our attack against  models on the \texttt{ImageNet} task while we summarize our main contributions below:

\begin{itemize}[itemsep=1pt,parsep=1pt,topsep=0pt]
    \item We formulate a new sparse attack---\SpaSB---in the score-based setting. The algorithm exploits the knowledge of model output scores and our intuitions on: \textbf{\textit{i)}}~learning influential pixel information from historical pixel manipulations; and \textbf{\textit{ii)}}~informed selection of pixel perturbations based on pixel dissimilarity between our search space prior and a source image to accelerate the search for a \textit{sparse} adversarial example. 
    
    \item As a \textit{first}, investigate the robustness of ViT and compare its relative robustness with ResNet models on the high-resolution dataset \texttt{Imagenet} under score-based sparse settings. 

    \item We demonstrate the significant query efficiency of our algorithm over the state-of-the-art counterpart in different datasets, against various deep learning models as well as defense mechanisms and {\small\textsf{Google Cloud Vision}} in terms of ASR \& sparsity under 10K query budgets. 
\end{itemize}

% ====================================================================
\section{Related Work}
\label{sec:bakcground and related work}
% ====================================================================
\vspace{-3mm}
\noindent\textbf{Non-Sparse (Dense) Attacks ($l_2, l_\infty$).~}Extensive past works studied dense attacks in white-box~\citep{Goodfellow2014, Madry2017, Carlini2017, Dong2018, Wong2019, Xu2020} and black-box settings~\citep{Chen2017, Tu2019, Liu2019,Ilyas2019, Andriushchenko2020,Shukla2021,Vo2022RamBo}. Due to non-differentiable, high-dimensional and mixed (continuous \& discrete) search space encountered in sparse settings, adopting these methods is non-trivial (see analysis in \textbf{Appendix}~\ref{apdx:Decision-Based and l0 Adapted Attacks}). Recent work has explored sparse attacks in white-box settings~\citep{Papernot2016,Modas2019,Croce2019,Fan2020,Dong2020, Zhu2021}. Here we mainly review \textit{sparse} attacks in \textit{black-box} settings but compare with a \textit{white-box sparse attack} for interest in Section~\ref{subsec:Decision-Based and l0 Adapted Attacks}.

\noindent\textbf{\textit{Decision}-based Sparse Attacks ($l_0$).~}
Only few recent studies, \textsc{Pointwise}~\citep{Schott2019} and \SpaEvoAtt~\citep{Vo2022}, have tackled the difficult problem of sparse attacks in decision-based settings. The fundamental difference between decision-based and score-based settings is the output information (labels vs scores) and the \textbf{\textit{need}} for a target class image sample in decision-based algorithms. The label information hinders direct optimization from output information. So, decision-based sparse attacks rely on an image from a target class (targeted attacks) and gradient-free methods. This leads to a different set of problem formulations. We study and demonstrate that sparse attacks formulated for \textit{decision}-based settings do not lead to query-efficient attacks in score-based settings in Section~\ref{subsec:Decision-Based and l0 Adapted Attacks}.

\noindent\textbf{\textit{Score}-based Sparse Attacks ($l_0$).~}A score-based setting seemingly provides more information than a decision-based setting. But, the first attack formulations~\citep{Narodytska2017, Zhao2019,Croce2019} suffer from prohibitive computational costs (low query efficiency) and do not scale to high-resolution datasets~\ie \texttt{ImageNet}. The recent \SparseRS random search algorithm in~\citep{Croce2022} reports \textit{the} state-of-the-art, query-efficient, sparse attack and is a significant advance. But, large query budgets are still required to achieve low sparsity on high resolution tasks such as \texttt{ImageNet} in the more difficult targeted attacks. 

\section{Proposed Method}
\label{sec:Proposed Method}
We focus on exploring adversarial attacks in the context of score-based and sparse settings.  First, we present the general problem formulation for sparse adversarial attacks. Let $\boldsymbol{x} \in [0,1]^{c \times w \times h}$ be a normalized source image, where $c$ is the number of channels and $w,~h$ is the width and height of the image and $y$ is its ground truth label---the \textit{source class}. Let $f(\boldsymbol{x})$ denote a vector of all class probabilities---softmax scores---from a victim model and $f(r|\boldsymbol{x})$ denote the probability of class $r$. 
An attacker aims to search for an adversarial example ${{\boldsymbol{\Tilde{x}}}} \in [0,1]^{c \times w \times h}$ such that ${\boldsymbol{\Tilde{x}}}$ can be misclassified by the victim model (\textit{untargeted} setting) or classified as a target class $y_\text{target}$ (\textit{targeted} setting). Formally, in a targeted setting, for a given $\boldsymbol{x}$, a sparse attack aiming to search for the best adversarial example $\boldsymbol{x^*}$ can be formulated as a constrained combinatorial optimization problem:
\begin{equation}
\label{eq:problem formulation}
%\begin{aligned}
\boldsymbol{x^{*}} = \argmin_{{\boldsymbol{\Tilde{x}}}} \, L(f({\boldsymbol{\Tilde{x}}}),y_\text{target}) ~
\textrm{s.t.} ~ \|\boldsymbol{{\boldsymbol{x}-\boldsymbol{\Tilde{x}}}}\|_0 \leq B\,, \vspace{-2mm}
%\end{aligned}
\end{equation}

where $\|\|_0$  is the $l_0$ norm denoting the number of perturbed pixels, ${B}$ denotes a budget of perturbed pixels and $L$ denotes  the loss function of the victim model $f$'s predictions. This loss may be different from the training loss and remains unknown to the attacker. In practice, we adopt the loss functions in~\citep{Croce2022}, particularly \textit{cross-entropy loss} in targeted settings and \textit{margin loss} in untargeted settings. 
The problem with Equation~\ref{eq:problem formulation} is the large search space as we need to search colors, float numbers in $[0,1]$, for perturbing some optimal combination of pixels in the source image $\boldsymbol{{x}}$.

\subsection{New Problem Formulation to Facilitate a Solution}
\label{sec:Problem Formulation}
Sparse attacks aim to search for the \textit{positions} and \textit{color values} of perturbed pixels; for a normalized image, the color value of each channel of a pixel---RGB color value---can be a float number in $[0,1]$. Consequently, the search space is enormous. Instead of searching in the mixed (discrete and continuous), high-dimensional search space, we consider turning the mixed search space problem into a lower-dimensional, discrete search space problem. Subsequently, we propose a formulation that will aid the development of a new solution to the combinatorial search problem.

\noindent\textbf{Proposed Lower Dimensional Search Space.~}We introduce a simple but effective perturbation scheme. We uniformly sample, at random, a color image $\boldsymbol{x'} \in \{0,~1\}^{c \times w \times h}$---which we call the \textit{synthetic color image}---to define the color of perturbed pixels in the source image $\boldsymbol{x}$. In this manner, each pixel is allowed to attain arbitrary values in $[0,1]$ for each color channel, but the dimensionality of the space is reduced to a discrete space of size $w\times h$. The resulting search space is eight times smaller than 
the perturbation scheme in \SparseRS~\citep{Croce2022} (see an analysis in \textbf{Appendix}~\ref{apdx:Search space reformulation}). Surprisingly, our proposal is %shown to be 
incredibly effective, particularly in high-resolution images such as \texttt{ImageNet} (we provide a comparative analysis with alternatives in \textbf{Appendix}~\ref{apdx:Randomly Initialize Synthetic Images}).

\noindent\textbf{Search Problem Over the Lower Dimensional Space.~}Despite the lower-dimensional nature of the search space, a combinatorial search problem persists. As a remedy, we propose changing the problem of finding $\boldsymbol{\Tilde{x}}$ to finding a binary matrix $\boldsymbol{u}$ for selecting pixels to perturb in $\boldsymbol{x}$ to construct an adversarial instance. To that end, we consider choosing a set of pixels in the given image $\boldsymbol{x}$ to be replaced by pixels from the synthetic color image $\boldsymbol{x}' \in \{0,1\}^{c \times w \times h}$. These pixels are determined by a binary matrix $\boldsymbol{u} \in \{0,1\}^{w \times h}$ where $u_{i,j~}=1$ indicates a pixel to be replaced. The adversarial image is then constructed as $\Tilde{\boldsymbol{x}}=\boldsymbol{u}\boldsymbol{x}'+(\mathbf{1}-\boldsymbol{u})\boldsymbol{x}$ where $\mathbf{1}$ denotes the matrix of all ones with dimensions of $\boldsymbol{u}$, and each element of $\boldsymbol{u}$ corresponds to one pixel of $\boldsymbol{x}$ with $c$ channels.
Consequently, manipulating each pixel of $\Tilde{\boldsymbol{x}}$ corresponds to manipulating an element in $\boldsymbol{u}$. Therefore, rather than solving Equation \ref{eq:problem formulation}, we consider the equivalent alternative (proof is shown in \textbf{Appendix}~\ref{apdx:reformulate the original problem}):
\begin{equation}
\label{eq:problem formulation with binary vector}
\boldsymbol{u^{*}} = \argmin_{{\boldsymbol{u}}} \, \ell(\boldsymbol{u}) \quad \textrm{s.t.} ~ \|\boldsymbol{u}\|_0 \leq B\,,
\end{equation}
where $\ell(\boldsymbol{u}) := L(f(\boldsymbol{u}\boldsymbol{x}'+(\mathbf{1}-\boldsymbol{u})\boldsymbol{x}),y_{\text{target}})$. Although the problem in \ref{eq:problem formulation with binary vector} is combinatorial in nature and does not have a polynomial time solution, the formulation facilitates the use of two simple intuitions to iteratively generate better solutions---i.e. sparse adversarial samples. 

\subsection{A Bayesian Framework for the $l_0$ Constrained Combinatorial Search}
\label{sec:Bayesian Formulation for learning a model}

\begin{figure}[htb]
    \begin{center}
        \includegraphics[scale=0.85]{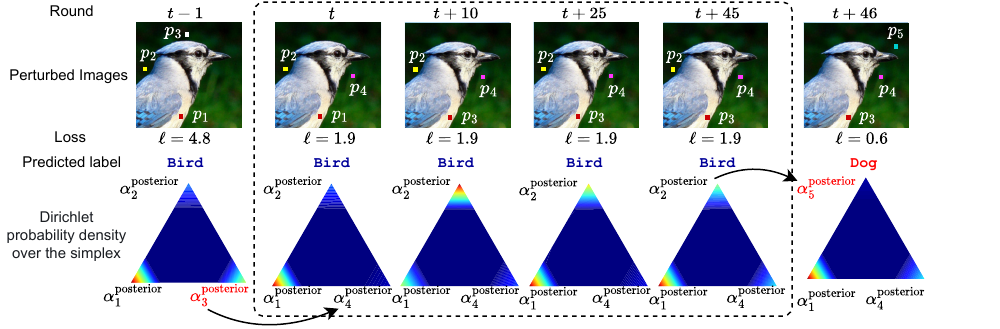}\vspace{-3mm}
        \caption[A {Sampling} and {Update} illustration.]{A \textbf{Sampling} and \textbf{Update} illustration. The attack aims to mislead a model into misclassifying a \texttt{Bird} image as \texttt{Dog}. Assuming that in round $t-1$, an adversarial instance is classified as \texttt{Bird} and loss $\ell=4.8$. We visualize \textit{three elements} of $\boldsymbol{\alpha}^{\text{posterior}}$ for simplicity. Let $\{p_1,p_2,p_3\}$ denote three perturbed pixels with corresponding posterior parameters $\{{\alpha}_1^{\text{posterior}},{\alpha}_2^{\text{posterior}},{\alpha}_3^{\text{posterior}}\}$. Assume that in round $t$, two pixels $p_1, p_2$ remain while $p_3$ is replaced by $p_4$ because a loss reduction is observed from 4.8 to 1.9. All $\{{\alpha}_1^{\text{posterior}},{\alpha}_2^{\text{posterior}},{\alpha}_3^{\text{posterior}}, {\alpha}_4^{\text{posterior}}\}$ are updated using Equation \ref{eq:posterior mean} but we visualize $\{{\alpha}_1^{\text{posterior}},{\alpha}_2^{\text{posterior}},{\alpha}_4^{\text{posterior}}\}$. Since ${\alpha}_4^{\text{posterior}}$ is new and has never been selected before, it is small in value (and represented using colder colors). From $t$ to $t+45$, while sampling and learning to find a better group of perturbed pixels,  $\boldsymbol{\alpha}^{\text{posterior}}$ is updated. Because $p_1$ has a high influence on the model's prediction (represented using warmer colors), it is more likely to remain, while $p_2, p_4$ are more likely to be selected for a replacement due to their lower impact on the model decision. In round $t+46$, pixel $p_2$ is replaced by $p_5$ because a loss reduction is observed from 1.9 to 0.6. Now, the predicted label is flipped from \texttt{Bird} to \texttt{Dog}.}
        \label{fig:Dirichlet prob density and alpha update}
    \end{center}
    \vspace{-3mm}
\end{figure}

It is clear that some pixels impart a more significant impact on the model decision than others. As such, given a binary matrix $\boldsymbol{u}$ with a set of selected elements---representing a candidate solution, we can expect some of these elements, if altered, to be more likely to result in an increase in the loss $\ell(\boldsymbol{u})$. Then, our assumption is that some selected elements must be \textit{hard to manipulate} to reduce the loss, and as such, should be unaltered. Retaining these selected elements is more likely to circumvent a bad solution successfully. \textit{In other words, these selected elements may significantly influence the model's decision and are worth keeping. In contrast to a stochastic search for influential pixels, 
we consider learning the influence of each element based on the history of pixel manipulations.} 

The influence of these elements can be modeled probabilistically, with the more influential elements attaining higher probabilities. To this end, we consider a categorical distribution parameterized by $\vtheta$, because we aim to select multiple elements and this is equivalent to multiple draws of one of many possible categories. It then follows to consider a Bayesian formulation to learn $\vtheta$ similar to \citet{infinite2017}. 
We adopt a general Bayesian framework and \textbf{\textit{design the new components and approximations}} needed to learn $\vtheta$. 
We can expect a new solution, $\boldsymbol{u^{(t)}}$, generated according to $\vtheta$ to more likely outweigh the current solution and guide the future candidate solution towards a pixel combination that more effectively minimizing the loss $\ell(\boldsymbol{u})$. Next, we describe these components and defer the algorithm we have designed, incorporating these components to Section~\ref{sec:Framework for Sparse Attacks}.

\noindent\textbf{Prior.~}In Bayesian statistics, the conjugate prior distribution of the categorical distribution is the Dirichlet distribution. Thus, we give $\vtheta$ a prior distribution defined by a Dirichlet distribution with the concentration parameter $\boldsymbol{\alpha}$ as $\operatorname{P}(\vtheta; \boldsymbol{\alpha}) :=  \operatorname{Dir}(\boldsymbol{\alpha}) \label{eq:prior distribution of parameter}.$

\noindent\textbf{Sampling $\boldsymbol{u}^{(t)}$.~} For $t>0$, given a solution---binary matrix $\boldsymbol{u}^{(t-1)}$---and $\vtheta^{(t)}$, we aim to: i)~select and preserve highly influential selected elements (Equation~\ref{eq:define v}); and ii)~draw new elements from unselected elements (Equation \ref{eq:define q}), conditioned upon $\boldsymbol{u}^{(t-1)}=\boldsymbol{1}$ and $\boldsymbol{u}^{(t-1)}=\boldsymbol{0}$, respectively, to jointly yield a new solution $\boldsymbol{u}^{(t)}$ (Equation \ref{eq:define u}). Concretely, we can express this process as follows:
{\begin{align}
\vv^{(t)}_1\ldots,\vv^{(t)}_b &\sim \operatorname{Cat} (\vv\mid\vtheta^{(t)}, \boldsymbol{u}^{(t-1)}=\mathbf{1}), \label{eq:define v}
\\
\vq^{(t)}_{1},\ldots,\vq^{(t)}_{B-b} &\sim \operatorname{Cat}(\vq\mid\vtheta^{(t)},\boldsymbol{u}^{(t-1)}=\mathbf{0}), \label{eq:define q}
\\
\boldsymbol{u}^{(t)} &= [\vee_{k=1}^b \vv_k^{(t)}] \vee [\vee_{r=1}^{B-b} \vq_k^{(t)} ]\,. \label{eq:define u} 
\end{align}}
Here $\vv^{(t)}_k, \vq^{(t)}_r \in \{0,1\}^{w \times h}$, $B$ denotes a total number of selected elements (a perturbation budget), $b$ denotes the number of selected elements remaining unchanged, and $\vee$ denotes logical \textit{OR} operator.

\noindent\textbf{Updating $\boldsymbol{\theta^{(t)}}$ (Using Our Proposed Likelihood).~} Finding the exact solution for the underlying parameters $\vtheta^{(t)}$ of the categorical distribution in Equation \ref{eq:define v} and Equation \ref{eq:define q} to increase the likelihood of yielding a better solution for $\boldsymbol{u}^{(t)}$ in Equation~\ref{eq:define u} is often intractable. Our approach is to find an estimate of $\vtheta^{(t)}$ by obtaining the expectation of the posterior distribution of the parameter, which is learned and updated over time through Bayesian inference. Notably, since the prior distribution of the parameter is a Dirichlet, which is the conjugate prior of the categorical (\ie distribution of $\boldsymbol{u}$), the posterior of the parameter is also Dirichlet. Formally, at each step $t>0$, updating the posterior and $\boldsymbol{\theta^{(t)}}$ is formulated as follows: 
\begin{align}
{\alpha}^{\text{posterior}}_{i,j}  &= {\alpha}^{\text{prior}}_{i,j}+ {s}_{i,j}^{(t)}%(\boldsymbol{u}^{(t)}, \ell^{(t)})
\label{eq:posterior mean}\\
\operatorname{P}(\vtheta \mid \boldsymbol{\alpha}, \boldsymbol{u}^{(t-1)}, \ell^{(t-1)}) :&=  \operatorname{Dir}(\boldsymbol{{\alpha}}^{\text{posterior}}) \label{eq:updated Dir}\\
\vtheta^{(t)} &= \E_{\vtheta \sim \operatorname{P}(\vtheta \mid \boldsymbol{\alpha}, \boldsymbol{u}^{(t-1)}, \ell^{(t-1)})}[\vtheta], \label{eq:define and update theta}
\end{align}
where $\boldsymbol{\alpha}^{\text{prior}}=\boldsymbol{\alpha}^{(0)}$ is the initial concentration parameter, $\boldsymbol{\alpha}^{\text{posterior}}= \boldsymbol{\alpha}^{(t)}$ denotes the updated concentration parameter (illustration in Figure \ref{fig:Dirichlet prob density and alpha update}) and ${s}^{(t)}_{i,j} =({(a_{i,j}^{(t))}+z)}/{(n_{i,j}^{(t)}+z)}) -1 $. Here, $z$ is a small constant (\ie~0.01) to ensure that the nominator and denominator are always non-zero (this smoothing technique is applied since the nominator and denominator can be zero when ``never" manipulated pixels are selected), $a_{i,j}^{(t)}$ is the accumulation of altered pixel $i,j$ (\ie $u^{(t)}_{i,j}=0$ and $u^{(t-1)}_{i,j}=1$) when it leads to an increase in the loss, \ie $\ell^{(t)} \geq \ell^{(t-1)}$, and $n_{i,j}^{(t)}$ is the accumulation of selected pixel $i,j$ in the mask $\boldsymbol{u}^{(t)}$. Formally, $a_{i,j}^{(t)}$ and $n_{i,j}^{(t)}$ can be updated as follows:%the following: 
\begin{flalign}
& a_{i,j}^{(t)} = \begin{cases} \label{eq:define s}
                     a_{i,j}^{(t-1)} + 1 & \text{if} \quad \ell^{t} \geq \ell^{(t-1)} \land u_{i,j}^{(t)}=1 \land u_{i,j}^{(t-1)}=0 \\
                     a_{i,j}^{(t-1)} & \text{otherwise}
                     \end{cases}\\
& n_{i,j}^{(t)} = \begin{cases} \label{eq:define n}
                     n_{i,j}^{(t-1)} + 1 & \text{if} \quad u_{i,j}^{(t)}=1 \lor u_{i,j}^{(t-1)}=1\\
                     n_{i,j}^{(t-1)} & \text{otherwise}
                     \end{cases}
\end{flalign}

\begin{figure}[t!]
    \begin{center}
        \includegraphics[scale=0.7]{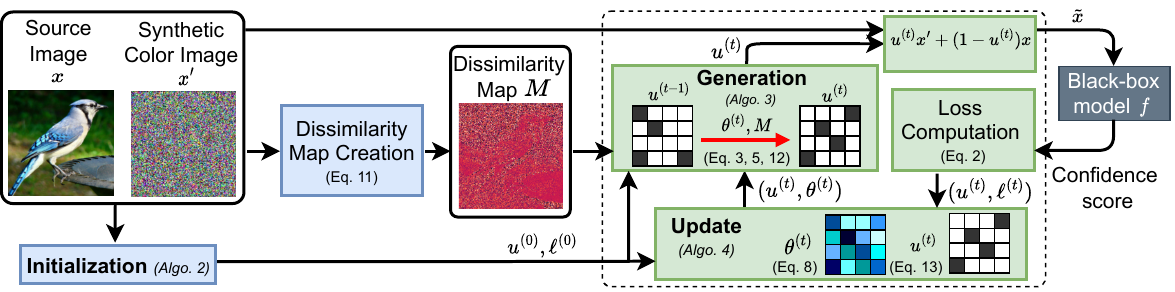}
        \caption[\SpaSB algorithm]{\SpaSB algorithm (Algo.~\ref{algo:main}). We aim to search for a set of pixels to replace in the source image $x$ by corresponding pixels in a synthetic color image $x^\prime$. In the solution, \textit{binary matrix} $\boldsymbol{u^{(t)}}$, \textit{white} and black colors denote \textit{replaced} and non-replaced pixels of the source image, respectively. Instead of a stochastic search, we employ our \textbf{Bayesian framework} in \textbf{$\S$\ref{sec:Bayesian Formulation for learning a model}}. \textbf{First}, we aim to retain useful elements in the solution $\boldsymbol{u^{(t)}}$ by learning from historical pixel manipulations. For this, we explore and \textbf{\textit{learn}} the influence of selected elements by capturing it in the model $\boldsymbol{\theta}$ using our general Bayesian framework in $\S$\ref{sec:Bayesian Formulation for learning a model}---darker colors illustrate the higher influence of selected elements %, to seek a better set of replacement pixels
        (Algo.~\ref{algo:Update}). \textbf{Second}, we %select and 
        \textit{generate} new pixel perturbations based on $\boldsymbol{\theta}$ with the \textbf{\textit{intuition}} that a larger pixel dissimilarity $M$ between our search space $x^\prime$ and a source image can possibly move the adversarial to the decision boundary faster and accelerate the search (Algo.~\ref{algo:generation}).}  
        \label{fig:main algorithm diagram}
    \end{center}
    \vspace{-4mm}
\end{figure}

\begin{algorithm}[t!]
    \SetKwInOut{KwIn}{Input}
    \DontPrintSemicolon
    \KwIn{source image $\boldsymbol{x}$, synthetic color image $\boldsymbol{x'}$, source label $y$, target label $y_\text{target}$, model $f$\; 
    \quad\quad ~query limit $T$, scheduler parameters $m_1,m_2$, initial changing rate $\lambda_0$\;
    \quad\quad ~perturbation budget $B$, a number of initial samples $N$, concentration parameters $\boldsymbol{\alpha}^{\text{prior}}$\;} 
    Create Dissimilarity Map $\mM$ using Equation \ref{eq:Deviation heat map}\; 
    $ \boldsymbol{u}^{(0)},\ell^{(0)} \leftarrow \textsc{Initialization} (\boldsymbol{x},\boldsymbol{x'},y,y_{\text{target}}, N, B, f)$ \;
    $t\leftarrow 1,~ \boldsymbol{a}^{(0)} \leftarrow \mathbf{0}, ~ \vn^{(0)} \gets \boldsymbol{u}^{(0)}$ \;
    Calculate $\boldsymbol{\theta}^{(0)}$ using $\boldsymbol{\alpha}^{\text{prior}}$  and Equation \ref{eq:define and update theta}\;
    \While{$t < T$ and $y^{(t)} \neq y_{\text{target}}$}{
        $\lambda^{(t)} \leftarrow \lambda_0(t^{m_1}+m_2^t) $ \;
      
        $\boldsymbol{u}^{(t)} \leftarrow \textsc{Generation}(\boldsymbol{\theta}^{(t)}, \mM, \boldsymbol{u}^{(t-1)},\lambda^{(t)})$ \;
        $\ell^{(t)} \leftarrow L(f(\boldsymbol{u}^{(t)}\boldsymbol{x}'+(\mathbf{1}-\boldsymbol{u}^{(t)})\boldsymbol{x}),y_\text{target}) $ \;
        $y^{(t)} \leftarrow \argmax_{r} f(r|\boldsymbol{u}^{(t)}\boldsymbol{x}'+(\mathbf{1}-\boldsymbol{u}^{(t)})\boldsymbol{x})$ \;
        
        $\boldsymbol{u}^{(t)},\ell^{(t)},\boldsymbol{\theta}^{(t)},\boldsymbol{a}^{(t)}, \boldsymbol{n}^{(t)}\leftarrow$ \textsc{Update} ($\boldsymbol{u}^{(t)},\ell^{(t)},\boldsymbol{u}^{(t-1)},\ell^{(t-1)}, \boldsymbol{a}^{(t)}, \boldsymbol{n}^{(t)}$)\;
        $t \leftarrow t+1$\;
        }
    \KwRet{$\boldsymbol{u}^{(t)}$}
    \caption{\SpaSB}
    \label{algo:main}
\end{algorithm}
\vspace{-7mm}

\vspace{2mm}
\subsection{Sparse Attack Algorithm Formulation With Our Bayesian Framework} 
\label{sec:Framework for Sparse Attacks}
Using the Bayesian framework for $l_0$ constrained combinatorial search in Section~\ref{sec:Bayesian Formulation for learning a model}, we devise our sparse attack (Algorithm \ref{algo:main}) illustrated in Figure \ref{fig:main algorithm diagram} and discuss it in detail as follows:

\noindent\textbf{Initialization~}(Algorithm \ref{algo:initialization}). Given a perturbation budget $B$ and a zero-initialized matrix $\boldsymbol{u}$, $N$ first solutions are generated by uniformly altering $B$ elements of $\boldsymbol{u}$ to 1 at random. The initial $\boldsymbol{u}^{(0)}$ is the solution incurring the lowest loss $\ell^{(0)}$. $\boldsymbol{\theta}^{(0)}$ is the expectation of $\boldsymbol{\alpha}^{\text{prior}}$ presented in Appendix \ref{apdx:the choice of alpha_prior}.

\noindent\textbf{Generation~} (Algorithm \ref{algo:generation}).
It is necessary here to balance exploration versus exploitation, as in other optimization methods. Initially, to explore the search space, we aim to manipulate a large number of selected elements. When approaching an optimal solution, we aim at exploitation to search for a solution in a region nearby a given solution and thus alter a small number of selected elements. Therefore, we use the combination of power and step decay schedulers to regulate a number of selected elements altered in round $t$. This scheduler is formulated as $\lambda_t = \lambda_0(t^{m_1} + m_2^t)$, where $\lambda_0$ is an initial changing rate, $m_1, m_2$ are power and step decay parameters respectively. Concretely, we define a number of selected elements remaining unchanged as $b=\lceil (1-\lambda_t)B \rceil$. 

Given a prior concentration parameter $\boldsymbol{\alpha}^{\text{prior}}$, to generate a new solution in round $t$, we first find $\boldsymbol{\alpha}^{\text{posterior}}$ as in Equation \ref{eq:posterior mean} and estimate $\vtheta^{(t)}$ as in Equation \ref{eq:define and update theta}. We then generate $\vv^{(t)}_k$ and $\vq^{(t)}_r$ as in Equation \ref{eq:define v} and Equation \ref{eq:define q}, respectively. A new solution $\boldsymbol{u}^{(t)}$ can be then formed as in Equation \ref{eq:define u}. Nonetheless, the naive approach of sampling $\vq^{(t)}_r$ as in Equation \ref{eq:define q} is ineffective and achieves a low performance at low levels of sparsity as shown in Appendix \ref{apdx:With vs Without Leveraging Prior Knowledge}. 
When altering unselected elements that are equivalent to replacing non-perturbed pixels in the source image with their corresponding pixels from the synthetic color image, the adversarial instance moves away from the source image by a distance. At a low sparsity level, since a small fraction of unselected elements are altered, the adversarial instance is able to take small steps toward the decision boundary between the source and target class. To mitigate this problem (taking inspiration from \citep{Brunner2019}) we employ a prior knowledge of the \textit{pixel dissimilarity} between the source image and the synthetic color image. Our intuition is that larger pixel dissimilarities lead to larger steps. As such, it is possible that altering unselected elements with a large pixel dissimilarity moves the adversarial instance to the decision boundary faster and accelerates optimization. The pixel dissimilarity is captured by a dissimilarity map $\mM$ as follows:
%--------------------------------------------------------------
\begin{equation}
\label{eq:Deviation heat map}
\mM = \frac{\sum_{c=0}^2 |\boldsymbol{x}_{c} - \boldsymbol{x}'_{c}|}{3},
\end{equation}
%--------------------------------------------------------------
where $c$ denotes a channel of a pixel. In practice, to incorporate $\mM$ into the step of sampling $\vq^{(t)}_r$, Equation \ref{eq:define q} is changed to the following: 
\begin{equation}
\vq^{(t)}_{1},\ldots,\vq^{(t)}_{B-b} \sim \operatorname{Cat}(\vq\mid\vtheta^{(t)}\mM,\boldsymbol{u}^{(t-1)}=\mathbf{0}) \label{eq:define q with M}
\end{equation}
%------------------------------------------------------------------
% 3. Loss computation
%------------------------------------------------------------------
\noindent\textbf{Update~}(Algorithm \ref{algo:Update}).
The generated solution $\boldsymbol{u}^{(t)}$ is associated with a loss $\ell^{(t)}$ given by the loss function in Equation \ref{eq:problem formulation with binary vector}. 
%------------------------------------------------------------------
% 4. Model update and information aggregation
%------------------------------------------------------------------
This is then used to update $\boldsymbol{\alpha}^{\text{posterior}}$ (Equation \ref{eq:posterior mean} and illustration in Figure \ref{fig:Dirichlet prob density and alpha update}) and the accepted solution %for each round $t$ 
as the following: 
{\small\begin{flalign}
\boldsymbol{u}^{(t)} = \begin{cases}
            \boldsymbol{u}^{(t)} &\text{if} \quad \ell^{(t)}<\ell^{(t-1)} \\
            \boldsymbol{u}^{(t-1)} &\text{otherwise~}
            \end{cases}\vspace{-3mm}
\end{flalign}}
\section{Experiments and Evaluations} \label{sec:Evaluations}\label{subsec:experiment setting}

\noindent\textbf{Attacks and Datasets.~}For a comprehensive evaluation of \SpaSB, we compose of evaluation sets from \texttt{CIFAR-10}~\citep{Krizhevsky}, \texttt{STL-10}~\citep{Coates2011} and \texttt{ImageNet}~\citep{Deng2009}. For \texttt{CIFAR-10} and \texttt{STL-10}, we select 9,000 and 60,094 different pairs of the source image and target class respectively. For \texttt{ImageNet}, we randomly select 200 \textit{correctly} classified test images evenly distributed among 200 random classes from \texttt{ImageNet}. To reduce the computational burden of the evaluation tasks in the \textit{targeted} setting, five target classes are randomly chosen for each image. For attacks against defended models with Adversarial Training, we randomly select 500 \textit{correctly} classified test images evenly distributed among 500 random classes from \texttt{ImageNet}. We compare with the state-of-the-art \SparseRS~\citep{Croce2022}.
 
\noindent\textbf{Models.~}\label{model description} For convolution-based networks, we use models based on a state-of-the-art architecture---ResNet---\citep{He2016} including ResNet18 achieving 95.28\% test accuracy on \texttt{CIFAR-10}, ResNet-9 obtaining 83.5\% test accuracy on \texttt{STL-10}, pre-trained ResNet-50~\citep{Marcel2010} with a 76.15\% Top-1 test accuracy, pre-trained Stylized ImageNet ResNet-50---ResNet-50 (SIN)---with a 76.72\% Top-1 test accuracy \citep{Geirhos2019} on \texttt{ImageNet}. For the attention-based network, we use a pre-trained ViT-B/16 model achieving 77.91\% Top-1 test accuracy~\citep{Dosovitskiy2021}. For robust ResNet-50 models \footnote{https://github.com/MadryLab/robustness}, we use adversarially pre-trained $l_2/~l_\infty$ models ($l_2$-At and $l_\infty$-AT)~\citep{robustness2019} with 57.9$\%$ and 62.42$\%$ clean test accuracy respectively. 

\noindent\textbf{Evaluation Metrics.~}\label{metric description}We define a \textit{sparsity} metric as the number of perturbed pixels divided by the total pixels of an image. To evaluate the performance of an attack, we use \textit{Attack Success Rate} (ASR). A generated perturbation is successful if it can yield an adversarial example with sparsity \textit{below a given sparsity threshold}, then ASR is defined as \textit{the number of successful attacks over the entire evaluation set} at different sparsity thresholds. We measure the \textit{robustness} of a model by the accuracy of that model under an attack at different query limits and sparsity levels.

\subsection{Attack Transformers \& Convolutional Nets}\label{subsec: Attack against Conventional CNNs}

We carry out comprehensive experiments on \texttt{ImageNet} under the targeted setting to investigate sparse attacks against various Deep Learning models (standard ResNet-50, ResNet-50 (SIN) and ViT). The results for the targeted and untargeted setting are detailed in Appendix \ref{apdx:Attack against DL modes on ImageNet-targeted setting}. Additional results on \texttt{STL-10} and \texttt{CIFAR-10} are provided in Appendix \ref{apdx:Attack against CNNs on STL10} and \ref{apdx:Attack against CNNs on CIFAR-10} respectively.

\begin{figure}[htp]
    \begin{center}
        \includegraphics[scale=0.38]{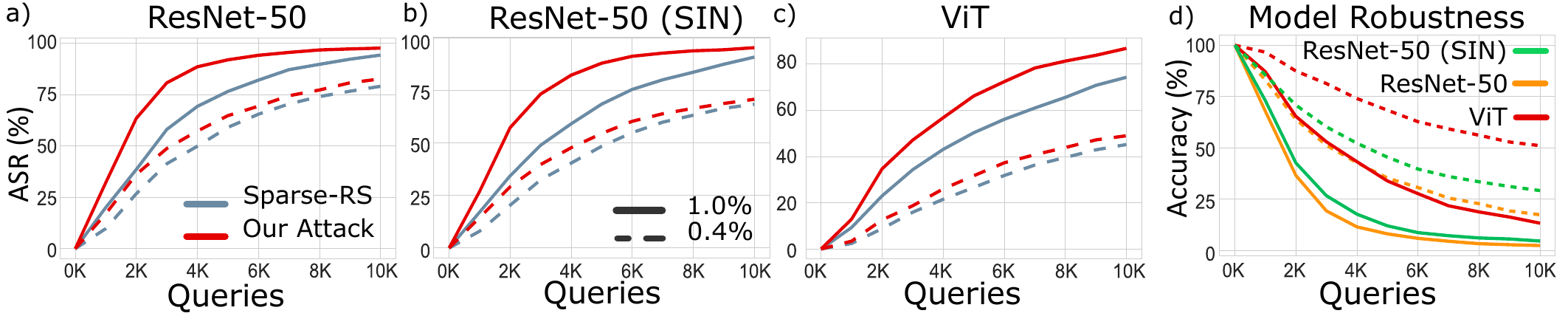}
        \caption{\textbf{Targeted setting} on $\texttt{ImageNet}$. a-c)~ASR of \SpaSB and \SparseRS against different models at sparsity levels of 0.4$\%$ (dashed lines) and 1.0$\%$ (solid lines); d)~Accuracy of different models against \SpaSB at sparsity levels (0.4$\%$ dash, 1.0$\%$ solid; in-between sparsity levels in Appendix~\ref{apdx:Attack against DL modes on ImageNet-targeted setting}).
        }
        \label{fig:imagenet_resnet_both}
    \end{center}
    \vspace{-6mm}
\end{figure}

\noindent\textbf{Convolutional-based Models.~}
Figure \ref{fig:imagenet_resnet_both}a and \ref{fig:imagenet_resnet_both}b  show that, at sparsity 0.4$\%$~($ \approx\frac{200}{224 \times 224}$), \SpaSB achieves slightly higher ASR than \SparseRS while at sparsity 1.0$\%$~($ \approx\frac{500}{224 \times 224}$), our attack significantly outweighs \SparseRS at different queries. Particularly, from 2K to 6K queries, \SpaSB obtains about 10$\%$ higher ASR than \SparseRS. Interestingly, with a small query budget of 6K queries, \SpaSB to achieve ASR higher than 90$\%$. 

\noindent\textbf{Attention-based Model.~}Figure \ref{fig:imagenet_resnet_both}c demonstrates that at sparsity of 0.4$\%$ \SpaSB achieves a marginally higher ASR than \SparseRS whereas at sparsity of 1.0$\%$ our attack demonstrates significantly better ASR than \SparseRS. At 1.0$\%$ sparsity and with query budgets above 2K, our method achieves roughly 10 $\%$ higher ASR than \SparseRS. Overall, our method consistently outperforms the \SparseRS in terms of ASR across different query budgets and sparsity levels.

\noindent\textbf{The Robustness of Transformer versus CNN.~}Figure \ref{fig:imagenet_resnet_both}d demonstrates the robustness of ResNet-50, ResNet-50 (SIN) and ViT models to adversarially sparse perturbation in the targeted settings. We observe that the performance of all three models degrades as expected. Although ResNet-50 (SIN) is more robust to several types of image corruptions than the standard ResNet-50 by far as shown in~\citep{Geirhos2019}, it is as vulnerable as its standard counterpart against sparse adversarial attacks. Interestingly, our results in Figure \ref{fig:imagenet_resnet_both}d illustrate that ViT is \textit{much less susceptible} than ResNet family against adversarially sparse perturbation. At the sparsity of 0.4$\%$ and 1.0 $\%$, the accuracy of ViT is pragmatically higher than both ResNet models under our attack across different queries. Interestingly, \SpaSB merely requires a \textit{small query budget of 4K} to degrade the accuracy of both ResNet models to the same accuracy of ViT at 10K queries. These findings can be explained that ViT's receptive field spans over the whole image \citep{Naseer2021} because some attention heads of ViT in the lower layers pay attention to the entire image \citep{Paul2022}. It is thus capable of enhancing relationships between various regions of the image and is harder to be evaded than convolutional-based models if a small subset of pixels is manipulated. 

%%%%%%%%%%%%%%%%%%%%%%%%%%%%%%%%%%%%%%%%%%%%%%%%%%%%%%%%%%%%%%%%%%
\subsection{Compare with Prior Decision-Based and $l_0$-Adapted Attack Algorithms}
\label{subsec:Decision-Based and l0 Adapted Attacks}%\ehsan{fix the figure ??}
\begin{wrapfigure}[]{l}{0.35\textwidth}
        \includegraphics[width=\linewidth]{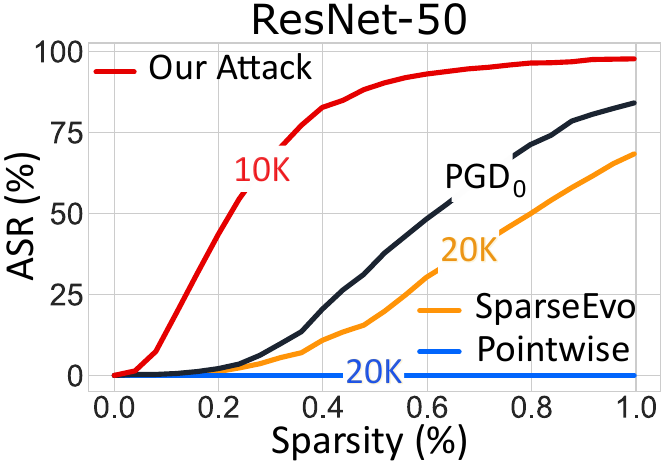}
        \caption{Targeted attacks on the \texttt{ImageNet} task against ResNet-50. ASR comparisons between \SpaSB and baselines: i)~\SpaEvoAtt and \textsc{Pointwise} (SOTA algorithms from \textbf{\textit{decision-based}} settings); ii)~\pgdlO (\textbf{\textit{whitebox}}).}
	\label{fig:sparesb versus adapted}
 \vspace{-4mm}
\end{wrapfigure}
In this section, we compare our method (10K queries) with baselines---\SpaEvoAtt  \citep{Vo2022}, Pointwise \citep{Schott2019} (both 20K queries) and \pgdlO \citep{Croce2019, Croce2022} (white-box)---in targeted settings. Figure \ref{fig:sparesb versus adapted} demonstrates that \SpaSB significantly outperforms \SpaEvoAtt and \pgdlO. For \SpaEvoAtt and Pointwise, this is expected because decision-based attacks and have only access to the hard label. For \pgdlO, it is surprised but understandable since in the $l_0$ project step, \pgdlO has to identify the minimum number of pixels required for projecting such that the perturbed image remains adversarial but to the best of our knowledge, there is no effective projection method to identify the pixels that can satisfy this projection constraint. \textit{Solving $l_0$ projection problem also lead to another NP-hard problem \citep{Modas2019, Dong2020} and hinders the adoption of dense attack algorithms to the $l_0$ constraint}. Moreover, the discrete nature of the $l_0$ ball impedes its amenability to continuous optimization \citep{Croce2022}. Additional results for $l_0$ adapted attacks on \texttt{CIFAR-10} are presented in Appendix \ref{apdx:Decision-Based and l0 Adapted Attacks}. 

%%%%%%%%%%%%%%%%%%%%%%%%%%%%%%%%%%%%%%%%%%%%%%%%%%%%%%%%%%%%%%%%%%
\subsection{Attack Defended Models} \label{subsec: Attack against defended models}
\textit{\SpaSB versus \SparseRS.~}In this section, we investigate the robustness of sparse attacks (with a budget of 5K queries) against adversarial training-based models using Projected Gradient Descent (PGD) proposed by~\citep{Madry2017}---highly effective defense mechanisms against adversarial attacks~\citep{Athalye2018} and Random Noise Defense (RND) \citep{Qin2021}---a recent defense method designed for black-box attacks. The robustness of the attacks is measured by the degraded accuracy of defended models under attacks at different sparsity levels. The stronger an attack is, the lower the accuracy of a defended model is. Table \ref{table:ASR against defense methods} shows that \SpaSB consistently outweighs \SparseRS against different defense methods and different sparsity levels. Additional results on \texttt{CIFAR-10} is provided in Appendix \ref{apdx:Evaluate Against SOTA Robust Models}.
\begin{table}[htp]
		\caption{Robustness comparison (lower $\downarrow$ is stronger) against undefended and defended models employing widely applied adversarial train regimes and the recent RND balckbox attack defence on the \texttt{ImageNet} task. Robustness is measured by the degraded accuracy of models under attacks at different sparsity levels.}
  \vspace{-4mm}
            \begin{center}
		\resizebox{1.0\linewidth}{!}{
		\begin{tabular}{c|cc|cc|cc|cc}
            \toprule
            \multirow{2}{*}{Sparsity}& \multicolumn{2}{c}{Undefended Model}&\multicolumn{2}{c}{$l_\infty$-AT} & \multicolumn{2}{c}{$l_2$-AT}&\multicolumn{2}{c}{RND}\\
            \cline{2-9} 
            &\SparseRS &\SpaSB &\SparseRS &\SpaSB & \SparseRS & \SpaSB& \SparseRS & \SpaSB \\ 
            \hline \hline
            {0.04$\%$}&{33.6$\%$} &\textbf{24.0$\%$} & {43.8}$\%$ & \textbf{42.2}$\%$ & 89.8$\%$ & \textbf{88.4}$\%$& 90.8${\%}$ & \textbf{85.0}${\%}$\\ 
            %\midrule
            {0.08$\%$} &{13.2$\%$}&\textbf{6.8$\%$}& {26.8}$\%$  & \textbf{24.4}$\%$& {81.2}${\%}$ &\textbf{79.2}${\%}$ & 82.2${\%}$ & \textbf{72.6}${\%}$\\
            %\midrule
            {0.12$\%$} &{7.6$\%$}&\textbf{2.6$\%$}& {19.0}$\%$ & \textbf{18.4}$\%$  & 75.8${\%}$ &\textbf{73.8}${\%}$& 73.6${\%}$ & \textbf{61.0}${\%}$\\ 
            %\midrule
            {0.16$\%$} &{5.2$\%$}& \textbf{1.0$\%$}& {16.6}$\%$ & \textbf{14.8}$\%$  & {71.4}${\%}$ & \textbf{69.2}${\%}$& 64.8${\%}$ & \textbf{51.4}${\%}$ \\
            %\midrule
            {0.2$\%$} &{4.6$\%$}&\textbf{1.0$\%$}& {12.2}$\%$ & \textbf{11.8}$\%$  & {68.4}${\%}$ & \textbf{66.4}${\%}$& 56.8${\%}$ & \textbf{42.6}${\%}$ \\
            \bottomrule 
            \end{tabular}
            }
            \end{center}
		\label{table:ASR against defense methods}
  \vspace{-4mm}
\end{table}

\textit{Undefended and Defended Models.~}The results in Table \ref{table:ASR against defense methods} shows the accuracy of undefended versus defended models against sparse attacks across different sparsity levels. In particular, under \SpaSB and sparsity of 0.2$\%$, the accuracy of ResNet-50 drops to 1$\%$ while $l_\infty$-AT model is able to obtain 11.8$\%$. However, $l_2$-AT model and RND strongly resist adversarially sparse perturbation and remains high accuracy around 66.4$\%$ and 42.6 $\%$ respectively. Therefore, $l_2$-AT model and RND are more robust than $l_\infty$-AT model to defense a model against sparse attacks. 

\begin{figure}[htp]
    \begin{center}
        \includegraphics[scale=0.36]{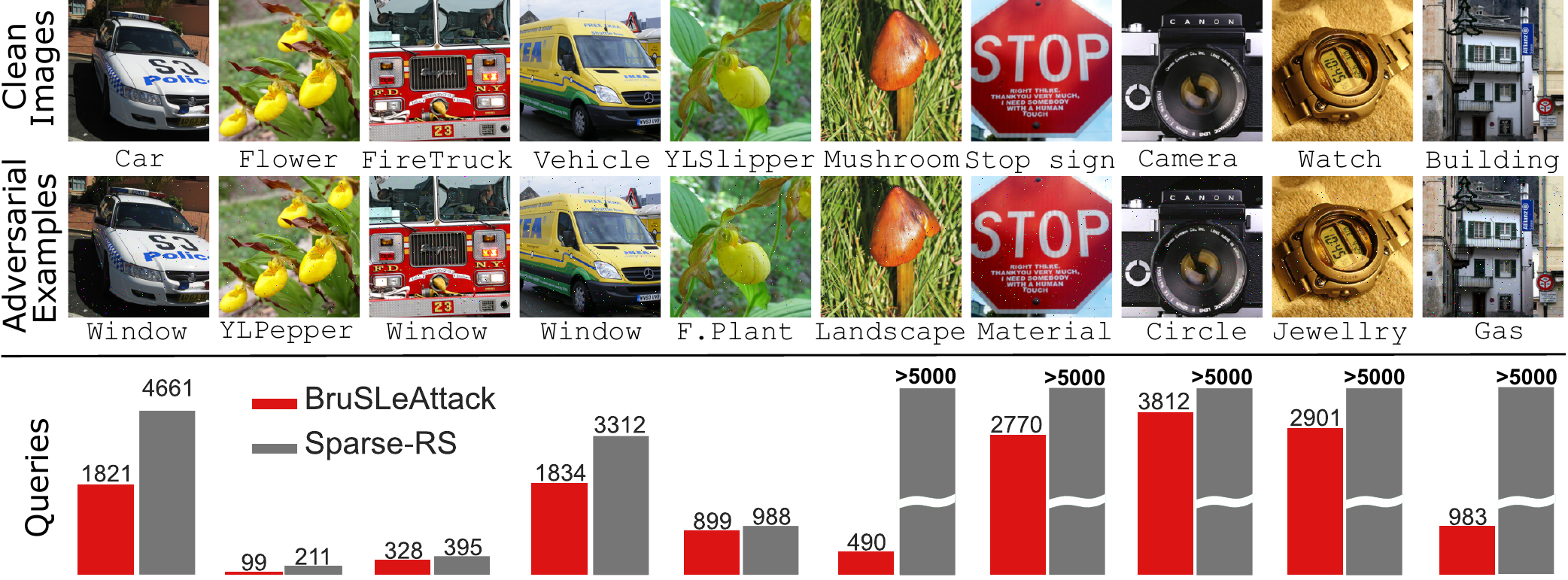}
        \caption{\textbf{Demonstration} of sparse attacks against GCV in targeted settings with a budget of 5K queries and sparsity of 0.5$\% \approx \frac{250}{224 \times 224}$. \SpaSB can yield adversarial examples for all clean images with less queries than \SparseRS while \SparseRS fails to yield adversarial examples for \texttt{Mushroom, Camera, Watch, $\&$ Building} images. Illustration on \textbf{GCV API} (online platform) is shown in Appendix \ref{apdx:Visualizations of Attack Against Google Cloud Vision}.}
        \label{fig:GCV-demo-query-bar}
    \end{center}
    \vspace{-5mm}
\end{figure}

\subsection{Attack Demonstration Against a Real-World System}
\label{subsec:Attack a Real-World System}
To illustrate the applicability and efficacy of \SpaSB against real-world systems, we attack the Google Cloud Vision (GCV) provided by Google. Attacking GCV is considerably challenging since \textit{1) the classifier returns partial observations of predicted scores with a varied length based on the input and 2) the scores are neither probabilities (softmax scores) nor logits} \citep{Ilyas2018, Guo2019}. To address these challenges, we employ the \textit{marginal loss} between the top label and the target label and successfully demonstrate our attack against GCV. With a budget of 5K queries and sparsity of 0.5$\%$, \SpaSB can craft a sparse adversarial example of all given images to mislead GCV whereas \SparseRS fails to attack four of them as shown in Figure \ref{fig:GCV-demo-query-bar}.

%===============================================================
\section{Conclusion}
%===============================================================
\vspace{-2mm}
In this paper, we propose a novel sparse attack---\SpaSB. We demonstrate that when attacking different Deep Learning models including undefended and defended models and in different datasets, \SpaSB consistently achieves better performance than the state-of-the-art method in terms of ASR at different query budgets. Tremendously, in a high-resolution dataset, our comprehensive experiments show that \SpaSB is remarkably query-efficient and reaches higher ASR than the current state-of-the-art sparse attack in score-based settings. 

\section*{Acknowledgements}
This work is supported in part by Google Cloud Research Credits Program (GCP19980904) and partially supported by the Australian Research Council (DP240103278). The attack is named in honor of Bruce Lee, a childhood hero---ours is a Q\underline{u}e\underline{r}y-Efficient \underline{S}core-Based B\underline{l}ack-Box Adv\underline{e}rsarial Attack built upon our proposed \underline{B}ayesian framework---\SpaSB. 

\bibliography{iclr2024_conference}
\bibliographystyle{iclr2024_conference}

\clearpage
\appendix

\section*{Overview of Materials in the Appendix}

\vspace{2mm}
We provide a brief overview of the extensive set of additional experimental results and findings in the Appendices that follows. Notably, given the significant computational resource required to mount blackbox attacks against models and our extensive additional experiments, we have employed \texttt{CIFAR-10} for the comparative studies. Importantly, our empirical results have already established the generalizability of our attack across CNN models, ViT models, three datasets and Google Cloud Vision.
\vspace{2mm}

\begin{enumerate}%[itemsep=3pt,parsep=3pt,topsep=3pt]
 \item \hl{A list of all notation used in the paper (Appendix \ref{apdx:notation table})}.
   \item Evaluation of score-based sparse attacks on \texttt{ImageNet} (targeted settings at sparsity levels between and including 0.4\% and 1.0\%; and untargeted settings) (Appendix~\ref{apdx:Attack against DL modes on ImageNet-targeted setting}).
    \item Evaluation of score-based sparse attacks on \texttt{STL-10} to demonstrate generalization %Details on hyperparameters 
    (Appendix~\ref{apdx:Attack against CNNs on STL10}).
    \item Evaluation of score-based sparse attacks on \texttt{CIFAR-10} demonstrate generalization (Appendix~\ref{apdx:Attack against CNNs on CIFAR-10}).
    \item Additional evaluation of attack algorithms adopted for sparse attacks ($l_0$ attacks) %Ablation study results and discussion 
    (Appendix~\ref{apdx:Decision-Based and l0 Adapted Attacks})
     \item Comparing \SpaSB and \SpaEvoAtt to supplement the results in Figure~\ref{fig:sparesb versus adapted} 
     (Appendix~\ref{apdx:Clarification about The Distinction})
     \item Demonstrating the impact of the Bayesian framework based search (Comparison with an adapted \SparseRS using our synthetic images. Notably, \textit{this addresses the feedback in the Meta Review}) (Appendix~\ref{apdx:sparseRSSynthetic})
    \item \hl{A Discussion Between \SpaSB (Adversarial Attack) and B3D (Black-box Backdoor Detection)~(Appendix~\ref{apdx:discussion between ours and B3D})}
    \item Additional evaluation of score-based sparse attacks against state of the art robust models from Robustbench 
    (Appendix~\ref{apdx:Evaluate Against SOTA Robust Models}).
    \item Proof of the optimization reformulation (Appendix~\ref{apdx:reformulate the original problem})
    \item An analysis of the search space reformulation and dimensionality reduction.
    (Appendix~\ref{apdx:Search space reformulation}).
    \item An analysis of different generation schemes for synthetic images we considered 
    (Appendix~\ref{apdx:Randomly Initialize Synthetic Images}).
    \item Study of \SpaSB performance under different random seeds 
    (Appendix~\ref{apdx:Study the performance variation}).
    \item An analysis of the effectiveness of the dissimilarity map employed in our proposed attack algorithm  
    (Appendix~\ref{apdx:With vs Without Leveraging Prior Knowledge}).
    \item Detailed information on the consistent set of hyper-parameters employed, initialization value for $\alpha^{prior}$ and computation resources used 
    (Appendix~\ref{apdx:hyper-parameters}).
    \item The notable performance invariance to hyper-parameter choices studies with \texttt{CIFA-10} and \texttt{ImageNet}  
    (Appendix~\ref{apdx:Hyper-parameters and scheme Impacts}).
    \item Additional study of employing different schedulers
    (Appendix~\ref{apdx:Different Schedulers}).
    \item Detailed implementation and pseudocodes of different components of \SpaSB
    (Appendix~\ref{apdx:Algorithm Pseudocodes}).
    \item Detailed information on the evaluation protocols \SpaSB (Appendix~\ref{apdx:Evaluation Protocol}).
    \item Visualizations of sparse attack against Google Cloud Vision
    (Appendix~\ref{apdx:Visualizations of Attack Against Google Cloud Vision}).
    \item Additional visualizations of dissimilarity maps and sparse adversarial examples
    (Appendix~\ref{apdx:Visualization of Sparse Adversarial Examples}).
     
\end{enumerate}
\vspace{2mm}
\clearpage

\section{Notation Table}
\label{apdx:notation table}
\hl{In this section, we list all notations in Table~\ref{table: notation table} to help the reader better understand the notations used in this paper.}

\begin{table}[htp]
\caption{\hl{Notations used in the paper.}}
            \begin{center}
		\begin{tabular}{c l c}
            \toprule
            \hl{Notation} & \hl{Description} \\ 
            \hline \hline
            \hl{$\boldsymbol{x}$} & \hl{Source image} \\
            \hl{$\boldsymbol{\Tilde{x}}$} & \hl{Synthetic color image} \\
            \hl{$y$} & \hl{Source class} \\
            \hl{$y_\text{target}$} & \hl{Target class} \\
            \hl{$f(\boldsymbol{x})$} & \hl{Softmax scores}\\ 
            \hl{$L(.) ~\text{or}~ \ell(.)$} & \hl{Loss function}\\ 
            \hl{$B$} & \hl{A budget of perturbed pixels}\\
            \hl{$b$} & \hl{A number of selected elements remaining unchanged}\\
            \hl{$\boldsymbol{u}^{(t)}$} & \hl{A binary matrix to determine perturbed and unperturbed pixels}\\
            \hl{$\boldsymbol{v}^{(t)}$} & \hl{A binary matrix to determine perturbed pixels remaining unchanged}\\
            \hl{$\boldsymbol{q}^{(t)}$} & \hl{A binary matrix to determine new pixels to be perturbed}\\
            \hl{$\boldsymbol{\alpha}^\text{prior}$} & \hl{An initial concentration parameter}\\
            \hl{$\boldsymbol{\alpha}^\text{posterior}$} & \hl{An updated concentration parameter}\\
            \hl{$\boldsymbol{\theta}$} & \hl{Parameter of Categorical distribution}\\ 
            \hl{$\text{Dir}(\boldsymbol{\alpha})$} & \hl{Dirichlet distribution}\\ 
            \hl{$\text{Cat}(\boldsymbol{\theta})$} & \hl{Categorical distribution}\\ 
            \hl{$\lambda_\text{0}$} & \hl{An initial changing rate}\\
            \hl{$m_1$} & \hl{A power decay parameter}\\
            \hl{$m_2$} & \hl{A step decay parameter}\\
            \hl{$\boldsymbol{M}$} & \hl{Dissimilarity Map}\\ 
            \hl{$w,~h,~c$} & \hl{Width, height and number of channels of an image}\\
            \bottomrule 
            \end{tabular}
            %}
            \end{center}
		\label{table: notation table}
  \vspace{-4mm}
\end{table}

\section{Sparse Attack Evaluations On \texttt{ImageNet}} \label{apdx:Attack against DL modes on ImageNet-targeted setting}
%%%%%%%%%%%%%%%%%%%%%%%%%%%%%%%%%%%%%%%%%%%%%%%%%%%%%%%%%%%%%%%%%%%%%%%
\begin{table}[htp]
\caption{ASR at different sparsity levels across different queries (higher is better). A comprehensive comparison among different attacks (\SparseRS and \SpaSB) against various Deep Learning models on \texttt{ImageNet} in the targeted setting.}
\label{table:targeted results}
\begin{center}
\resizebox{0.90\linewidth}{!}{
\begin{tabular}{c|cc|cc|cc}
\toprule
\multirow{2}{*}{Query} &\multicolumn{2}{c}{\texttt{ResNet-50}} & \multicolumn{2}{c}{\texttt{ResNet-50(SIN)}} & \multicolumn{2}{c}{\texttt{ViT}}\\
\cline{2-7} 
&\SparseRS &\SpaSB & \SparseRS & \SpaSB & \SparseRS & \SpaSB\\ \hline \hline
 %====================================================================================
\multicolumn{7}{c}{Sparsity = 0.4$\%$}\\
\cline{1-7}
{4000}& {49.9}$\%$  & \textbf{57.3}$\%$& {40.5}$\%$ & \textbf{47.8}$\boldsymbol{\%}$ & {21.5}$\boldsymbol{\%}$ &\textbf{26.0}$\boldsymbol{\%}$ \\
{6000} & {65.5}$\%$ & \textbf{69.4}$\%$  & {55.0}$\%$ & \textbf{60.4}$\boldsymbol{\%}$& 31.8$\boldsymbol{\%}$ &\textbf{37.3}$\boldsymbol{\%}$\\ 
{8000}& {74.1}$\%$ & \textbf{77.3}$\%$ & {63.3}$\%$& \textbf{66.6}$\boldsymbol{\%}$  & {39.6}$\boldsymbol{\%}$ & \textbf{43.9}$\boldsymbol{\%}$\\
{10000} & {79.1}$\%$ & \textbf{82.7}$\%$ & {68.5}$\%$ & \textbf{70.9}${\%}$ & {45.2}$\boldsymbol{\%}$ & \textbf{49.0}$\boldsymbol{\%}$\\
%====================================================================================
\midrule
\multicolumn{7}{c}{Sparsity = 0.6$\%$}\\
\cline{1-7}
{4000}& {59.6}$\%$  & \textbf{75.1}$\%$& {49.7}$\%$ & \textbf{66.2}$\boldsymbol{\%}$ & {30.8}$\boldsymbol{\%}$ &\textbf{40.7}$\boldsymbol{\%}$ \\
%\midrule
%\cline{2-8} 
{6000} & {74.0}$\%$ & \textbf{86.3}$\%$  & {65.6}$\%$ & \textbf{77.8}$\boldsymbol{\%}$& 43.7$\boldsymbol{\%}$ &\textbf{52.0}$\boldsymbol{\%}$\\ 
%\midrule
{8000}& {85.0}$\%$ & \textbf{90.3}$\%$ & {77.6}$\%$& \textbf{83.4}$\boldsymbol{\%}$  & {52.2}$\boldsymbol{\%}$ & \textbf{61.0}$\boldsymbol{\%}$\\
{10000} & {90.9}$\%$ & \textbf{93.0}$\%$ & {84.3}$\%$ & \textbf{87.0}${\%}$ & {61.7}$\boldsymbol{\%}$ & \textbf{67.3}$\boldsymbol{\%}$\\
%====================================================================================
\midrule
\multicolumn{7}{c}{Sparsity = 0.8$\%$}\\
\cline{1-7}
{4000}& {65.8}$\%$ & \textbf{84.3}$\%$ & {56.3}$\%$ & \textbf{76.7}$\boldsymbol{\%}$& {38.2}$\boldsymbol{\%}$ & \textbf{49.4}$\boldsymbol{\%}$\\
%\midrule
%\cline{2-8} 
{6000} & {79.2} & \textbf{90.6}$\%$ & {71.1}$\%$ & \textbf{87.0}$\boldsymbol{\%}$& 50.2$\boldsymbol{\%}$ &\textbf{63.4}$\boldsymbol{\%}$\\ 
%\midrule
{8000} & {87.9}$\%$ & \textbf{94.3}$\%$ & {81.9}$\%$ & \textbf{91.0}${\%}$ & {60.0}$\boldsymbol{\%}$ & \textbf{72.2}$\boldsymbol{\%}$\\
{10000} & {93.4}$\%$ & \textbf{96.4}$\%$ & {89.6}$\%$ & \textbf{92.4}${\%}$ & {69.6}$\boldsymbol{\%}$ & \textbf{79.0}$\boldsymbol{\%}$\\
%====================================================================================
\midrule
\multicolumn{7}{c}{Sparsity = 1.0$\%$}\\
\cline{1-7}
{4000}& {69.3}$\%$ & \textbf{88.6}$\%$ & {59.2}$\%$ & \textbf{82.4}$\boldsymbol{\%}$& {43.1}$\boldsymbol{\%}$ & \textbf{56.8}$\boldsymbol{\%}$\\
{6000} & {82.1} & \textbf{94.2}$\%$ & {75.6}$\%$ & \textbf{91.4}$\boldsymbol{\%}$& 56.1$\boldsymbol{\%}$ &\textbf{72.4}$\boldsymbol{\%}$\\ 
{8000} & {89.8}$\%$ & \textbf{96.8}$\%$ & {83.8}$\%$ & \textbf{94.0}${\%}$ & {65.6}$\boldsymbol{\%}$ & \textbf{81.3}$\boldsymbol{\%}$\\
{10000} & {94.3}$\%$ & \textbf{97.7}$\%$ & {91.0}$\%$ & \textbf{95.5}${\%}$ & {74.3}$\boldsymbol{\%}$ & \textbf{86.8}$\boldsymbol{\%}$\\
\bottomrule 
\end{tabular}
}
\end{center}
\end{table}
%%%%%%%%%%%%%%%%%%%%%%%%%%%%%%%%%%%%%%%%%%%%%%%%%%%%%%%%%%%%%%
\textbf{Targeted Settings.~}Table \ref{table:targeted results} shows the detailed ASR results for sparse attacks on high-resolution dataset \texttt{ImageNet} in the targeted settings shown in Section \ref{subsec: Attack against Conventional CNNs}. The results illustrate that the proposed method is consistently better than \SparseRS across different sparsity levels from 0.4 $\%$ to 1.0 $\%$.
%%%%%%%%%%%%%%%%%%%%%%%%%%%%%%%%%%%%%%%%%%%%%%%%%%%%%%%%%%%%%%%%%%%%%%%

\begin{figure}[htp]
    \begin{center}
        \includegraphics[scale=0.37]{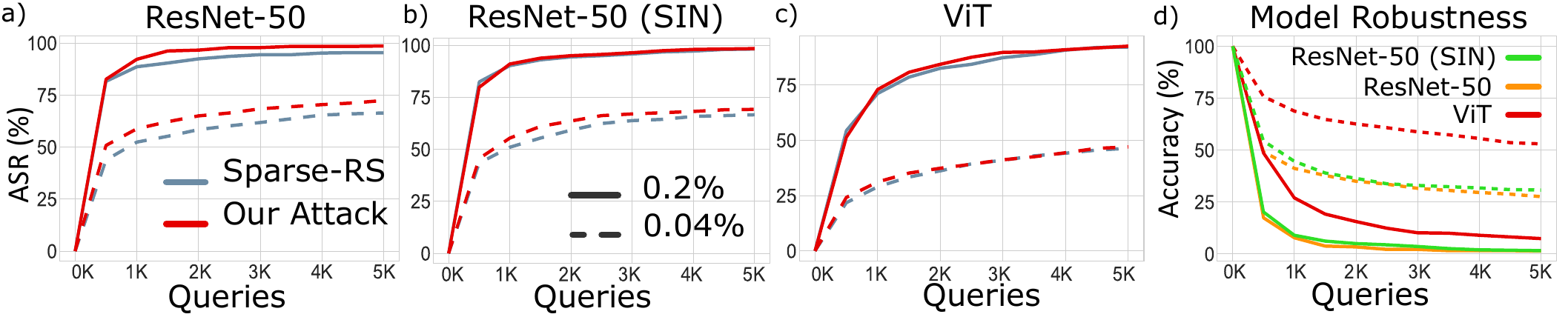}
        \caption{a-c)~\textbf{Untargetted Setting}. ASR versus the number of model queries against different Deep Learning models at sparsity levels (0.4$\%$, 1.0$\%$); d)~Accuracy versus the number of model queries for model robustness comparison against \SpaSB, in the untargeted setting and at sparsity levels (0.04$\%=\frac{40}{224 \times 224}$, 0.2$\%=\frac{100}{224 \times 224}$).}
        \label{fig:untargeted results}
    \end{center}
\end{figure}
% -----------------------------------------------------------------
% Untargeted Attack
% -----------------------------------------------------------------
\textbf{Untargeted Settings.~}In this section, we verify the performance of sparse attacks against different Deep Learning models including ResNet-50, ResNet-50 (SIN) and ViT models in the untargeted setting up to a 5K query budget. We use an evaluation set of 500 random pairs of an image and a target class to conduct this comprehensive experiment. Our results in Table \ref{table:untargeted results} and Table \ref{fig:untargeted results}a-c show that \SpaSB is marginally better than \SparseRS across different sparsity levels when attacking against ViT. For ResNet-50 and ResNet-50 (SIN), at lower sparsity or lower query limits, our proposed attack outperforms \SparseRS while at higher query budgets or higher sparsity levels, \SparseRS is able to obtain slightly lower ASR than our method. In general, \SpaSB consistently outperforms \SparseRS and only needs 1K queries and sparsity of 0.2$\%$ (100 pixels) to achieve above 90$\%$ ASR against both ResNet-50 and ResNet-50 (SIN).  
\begin{table}[htp]
\caption{ASR at different sparsity levels across different queries (higher is better). A comprehensive comparison among different attacks (\SparseRS and \SpaSB) and various DL models on \texttt{ImageNet} in the untargeted setting.}
\label{table:untargeted results}
\begin{center}
\resizebox{0.9\linewidth}{!}{
\begin{tabular}{c|cc|cc|cc}
\toprule
\multirow{2}{*}{Query} &\multicolumn{2}{c}{\texttt{ResNet-50}} & \multicolumn{2}{c}{\texttt{ResNet-50(SIN)}} & \multicolumn{2}{c}{\texttt{ViT}}\\
\cline{2-7} 
&\SparseRS &\SpaSB & \SparseRS & \SpaSB & \SparseRS & \SpaSB\\ \midrule
\multicolumn{7}{c}{Sparsity = 0.04$\%$}\\
\cline{1-7}
{1000} & 
{52.4}$\%$ & \textbf{58.8}$\%$ & {51.0}$\%$&  \textbf{55.4}$\boldsymbol{\%}$&29.0$\boldsymbol{\%}$ & \textbf{31.2}$\boldsymbol{\%}$ \\ 
{2000}& {58.4}$\%$  & \textbf{65.0}$\%$& {59.2}$\%$ & \textbf{63.6}$\boldsymbol{\%}$ & {36.2}$\boldsymbol{\%}$ &\textbf{37.4}$\boldsymbol{\%}$ \\
{3000} & {61.8}$\%$ & \textbf{68.4}$\%$  & {63.8}$\%$ & \textbf{67.0}$\boldsymbol{\%}$& 41.0$\boldsymbol{\%}$ &\textbf{41.2}$\boldsymbol{\%}$\\ 
{4000}& {65.4}$\%$ & \textbf{70.4}$\%$ & {65.8}$\%$& \textbf{68.2}$\boldsymbol{\%}$  & {44.2}$\boldsymbol{\%}$ & \textbf{44.4}$\boldsymbol{\%}$\\
{5000} & {66.4}$\%$ & \textbf{72.4}$\%$ & {66.6}$\%$ & \textbf{69.2}${\%}$ & {46.4}$\boldsymbol{\%}$ & \textbf{46.7}$\boldsymbol{\%}$\\
%====================================================================================
\midrule
\multicolumn{7}{c}{Sparsity = 0.08$\%$}\\
\cline{1-7}
{1000} & {72.8}$\%$ & \textbf{77.4}$\%$ & {73.8}$\%$&  \textbf{75.8}$\boldsymbol{\%}$&47.2$\boldsymbol{\%}$ & \textbf{50.6}$\boldsymbol{\%}$ \\ 
{2000}& {81.2}$\%$  & \textbf{86.8}$\%$& {80.4}$\%$ & \textbf{83.4}$\boldsymbol{\%}$ & {57.6}$\boldsymbol{\%}$ &\textbf{61.0}$\boldsymbol{\%}$ \\
{3000} & {84.6}$\%$ & \textbf{89}$\%$  & {84.4}$\%$ & \textbf{87.0}$\boldsymbol{\%}$& 64.2$\boldsymbol{\%}$ &\textbf{67.8}$\boldsymbol{\%}$\\ 
{4000}& {85.6}$\%$ & \textbf{90.4}$\%$ & {86.6}$\%$& \textbf{88.2}$\boldsymbol{\%}$  & {69.6}$\boldsymbol{\%}$ & \textbf{72.6}$\boldsymbol{\%}$\\
{5000} & {86.8}$\%$ & \textbf{90.8}$\%$ & {87.0}$\%$ & \textbf{88.6}${\%}$ & {72.6}$\boldsymbol{\%}$ & \textbf{74.6}$\boldsymbol{\%}$\\
%====================================================================================
\midrule
\multicolumn{7}{c}{Sparsity = 0.16$\%$}\\
\cline{1-7}
{1000} & {87.0}$\%$ & \textbf{89.4}$\%$& {87.6}$\%$ & \textbf{88.0}$\boldsymbol{\%}$ & 64.8$\boldsymbol{\%}$ &\textbf{68.6}$\boldsymbol{\%}$\\ 
{2000}& {90.8}$\%$ & \textbf{95.2}$\%$ & {92.0}$\%$ & \textbf{94.0}$\boldsymbol{\%}$& {78.4}$\boldsymbol{\%}$ & \textbf{81.4}$\boldsymbol{\%}$\\
{3000} & {93.4} & \textbf{96.8}$\%$ & {94.8}$\%$ & \textbf{95.6}$\boldsymbol{\%}$& 85.0$\boldsymbol{\%}$ &\textbf{86.4}$\boldsymbol{\%}$\\ 
{4000} & {94.4}$\%$ & \textbf{97.6}$\%$ & {96.2}$\%$ & \textbf{97.0}${\%}$ & {87.0}$\boldsymbol{\%}$ & \textbf{89.2}$\boldsymbol{\%}$\\
{5000} & {94.8}$\%$ & \textbf{98.4}$\%$ & {96.8}$\%$ & \textbf{97.4}${\%}$ & {89.8}$\boldsymbol{\%}$ & \textbf{90.0}$\boldsymbol{\%}$\\
%====================================================================================
\midrule
\multicolumn{7}{c}{Sparsity = 0.2$\%$}\\
\cline{1-7}
{1000} & {88.6}$\%$ & \textbf{92.2}$\%$& {90.2}$\%$ & \textbf{91.0}$\boldsymbol{\%}$ & 71.2$\boldsymbol{\%}$ &\textbf{73.0}$\boldsymbol{\%}$\\ 
{2000}& {92.4}$\%$ & \textbf{96.6}$\%$ & {94.4}$\%$ & \textbf{95.0}$\boldsymbol{\%}$& {82.6}$\boldsymbol{\%}$ & \textbf{84.4}$\boldsymbol{\%}$\\
{3000} & {94.4} & \textbf{97.8}$\%$ & {95.8}$\%$ & \textbf{96.4}$\boldsymbol{\%}$& 87.4$\boldsymbol{\%}$ &\textbf{89.8}$\boldsymbol{\%}$\\ 
{4000} & {95.2}$\%$ & \textbf{98.4}$\%$ & {97.2}$\%$ & \textbf{98.0}${\%}$ & {90.8}$\boldsymbol{\%}$ & \textbf{91.0}$\boldsymbol{\%}$\\
{5000} & {95.4}$\%$ & \textbf{98.6}$\%$ & {98.2}$\%$ & \textbf{98.4}${\%}$ & {92.2}$\boldsymbol{\%}$ & \textbf{92.6}$\boldsymbol{\%}$\\
\bottomrule 
\end{tabular}}
\end{center}
\end{table}

\textbf{Relative Robustness Comparison among Models.~} To compare the relative robustness of different models, we evaluate these models against our attack. Table \ref{table:untargeted results} and  Figure \ref{fig:untargeted results}d confirm our observations about relative robustness of ResNet-50 (SIN) to the standard ResNet-50 in the targeted setting (presented in Section \ref{subsec: Attack against Conventional CNNs}). It turns out that ResNet-50 (SIN) is as vulnerable as the standard ResNet-50 even though it is robust against various types of image distortion. Interestingly, ViT is more robust than its convolutional counterparts under sparse attack. Particularly, at sparsity of 0.2$\%$ and 2K queries, while the accuracy of both ResNet-50 and ResNet-50 (SIN) is down to about 5$\%$, ViT is still able to remain ASR around 15$\%$.

% -----------------------------------------------------------------
% CIFAR-10
% -----------------------------------------------------------------
\section{Sparse Attack Evaluations on \texttt{STL10} (Targeted Settings)} \label{apdx:Attack against CNNs on STL10}
We conduct more extensive experiments on \texttt{STL-10} in the targeted setting with all correctly classified images of the evaluation set (60,094 sample pairs and image size 96$\times$96). Table \ref{table:stl10-resnet9} provides a comprehensive comparison for different attacks across different sparsity levels ranging from 0.11$\%$ (10 pixels) to 0.54$\%$ (50 pixels). Particularly, with only 50 pixels, \SpaSB needs solely 3000 queries to achieve ASR beyond 92$\%$ whereas \SparseRS only reaches ASR of 89.64$\%$. 

\begin{table}[htp]
\caption{ASR (higher is better) at different sparsity levels in targeted settings. A comprehensive comparison between \SparseRS and \SpaSB against ResNet9 on a full evaluation set from \texttt{STL-10}.}
\label{table:stl10-resnet9}
\begin{center}
\resizebox{0.9\linewidth}{!}{
\begin{tabular}{c|cccc|cccc}
\toprule
{Methods}&{Q=1000} & {Q=2000} & {Q=3000} & {Q=4000}&{Q=1000} & {Q=2000} & {Q=3000} & {Q=4000}\\
%----------------------------------------------------------------
\hline
&\multicolumn{4}{c}{Sparsity = 0.22$\%$} & \multicolumn{4}{c}{Sparsity = 0.44$\%$}\\
\cline{1-9}
\SparseRS & 53.82$\%$ &{61.65$\%$} & {65.84$\%$}& {68.0$\%$}  &{73.34$\%$} & {81.47$\%$}& {85.24$\%$} & 87.49$\%$\\ 
\textbf{\SpaSB} & \textbf{57.69$\%$} & \textbf{65.05$\%$}& \textbf{68.8$\%$} & \textbf{71.22$\%$} & \textbf{78.21$\%$} & \textbf{85.03$\%$}& \textbf{88.31$\%$} & \textbf{90.26$\%$} \\
%----------------------------------------------------------------
\hline
&\multicolumn{4}{c}{Sparsity = 0.33$\%$}&\multicolumn{4}{c}{Sparsity = 0.54$\%$}\\
\cline{1-9}
\SparseRS & {65.6$\%$} & {74.0$\%$}& {78.0$\%$} & 80.65$\%$ & {78.66$\%$} & {86.31$\%$}& {89.64$\%$} & 91.61$\%$\\ %& 

\textbf{\SpaSB} & \textbf{70.27$\%$} & \textbf{77.55$\%$}& \textbf{81.16$\%$} & \textbf{83.42$\%$}
& \textbf{83.29$\%$} & \textbf{89.78$\%$}& \textbf{92.55$\%$} & \textbf{94.08$\%$} \\
%----------------------------------------------------------------
\bottomrule 
\end{tabular}}
\end{center}
\end{table}
% -----------------------------------------------------------------
% CIFAR-10
% -----------------------------------------------------------------
\section{Sparse Attack Evaluations on \texttt{CIFAR-10} (Targeted Settings)} \label{apdx:Attack against CNNs on CIFAR-10}
In this section, we conduct extensive experiments in the targeted setting to investigate the robustness of sparse attacks on an evaluation set of 9,000 pairs of an image and a target class from \texttt{CIFAR-10} (image size 32$\times$32). Sparsity levels range from 1.0$\%$ (10 pixels) to 3.9$\%$ (40 pixels). Table \ref{table:cifar10-resnet18} provides a comprehensive comparison of different attacks in the targeted setting. Particularly, with only 20 pixels (sparsity of 2.0 $\%$), \SpaSB needs solely 500 queries to achieve ASR beyond 90$\%$ whereas \SparseRS only reaches ASR of 89.21$\%$. Additionally, with only 300 queries, \SpaSB is able to reach above 95$\%$ of successfully crafting adversarial examples with solely 40 pixels. Overall, our attack consistently outperforms the \SparseRS in terms of ASR and this confirms our observations on \texttt{STL-10} and \texttt{ImageNet}.

\begin{table}[htp]
\begin{center}
\caption{ASR (higher is better) at different sparsity thresholds in the targeted setting. A comprehensive comparison among different attacks (\SparseRS and \SpaSB) against ResNet18 on an evaluation set of 9,000 pairs of an image and a target class from \texttt{CIFAR-10}.}
\label{table:cifar10-resnet18}
\resizebox{0.65\linewidth}{!}{
\begin{tabular}{c|ccccc}
\toprule
{Methods}&{Q=100} & {Q=200} & {Q=300} & {Q=400} & {Q=500}\\
\midrule
\multicolumn{6}{c}{Sparsity = 1.0$\%$}\\
\cline{1-6}
\SparseRS & {36.22}$\%$ & {50.6}$\%$& {58.17} $\%$& 62.59$\%$& {66.26$\%$}\\ 
\textbf{\SpaSB}  & \textbf{42.32$\%$} & \textbf{54.73$\%$}& \textbf{61.49$\%$} & \textbf{65.33$\%$} & \textbf{68.21$\%$} \\
%----------------------------------------------------------------
\midrule
\multicolumn{6}{c}{Sparsity = 2.0$\%$}\\
\cline{1-6}
\SparseRS & {60.51$\%$} & {76.1$\%$}& {83.13$\%$} & 86.89$\%$& 89.21$\%$\\ 
\textbf{\SpaSB} & \textbf{66.01$\%$} & \textbf{79.19$\%$}& \textbf{84.84$\%$} & \textbf{88.27$\%$} & \textbf{90.24$\%$} \\
%----------------------------------------------------------------
\midrule
\multicolumn{6}{c}{Sparsity = 2.9$\%$}\\
\cline{1-6}
\SparseRS & {71.29$\%$} & {85.67$\%$}& {91.21$\%$} & 94.28$\%$& 95.78$\%$\\ 
%& 
\textbf{\SpaSB} & \textbf{75.54$\%$} & \textbf{88.22$\%$}& \textbf{92.91$\%$} & \textbf{95.2$\%$} & \textbf{96.59$\%$} \\
%----------------------------------------------------------------
\midrule
\multicolumn{6}{c}{Sparsity = 3.9$\%$}\\
\cline{1-6}
\SparseRS & {75.91$\%$} & {90.21$\%$}& {94.78$\%$} & 96.97$\%$& 97.98$\%$\\ 
\textbf{\SpaSB} & \textbf{80.44$\%$} & \textbf{91.24$\%$}& \textbf{95.43$\%$} & \textbf{97.4$\%$} & \textbf{98.48$\%$} \\
\bottomrule 
\end{tabular}
}
\end{center}
\end{table}

\section{Comparing \SpaSB With Other Attacks Adapted for Score-Based Sparse Attacks For Additional Baselines} 
\label{apdx:Decision-Based and l0 Adapted Attacks}

\subsection{Additional Evaluations With Decision-Based Sparse Attack Methods}
\begin{figure}[htp]
    \begin{center}
        \includegraphics[scale=0.6]{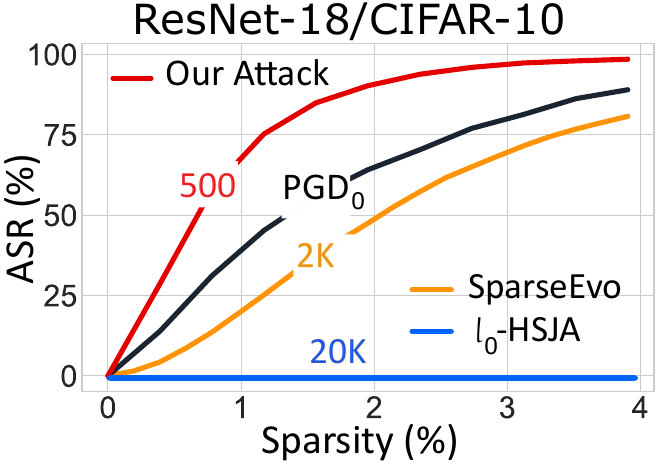}
        \caption{\textbf{Targeted attacks} on \texttt{CIFAR-10} against ResNet-18. ASR comparisons between \SpaSB and baselines i)~\SparseRS and adapted $\l_0$-HSJA (\textbf{\textit{decision-based}} settings); ii)~\pgdlO (\textbf{\textit{whitebox}}).}
	\label{fig:sparesb versus adapted attacks}
    \end{center}
\end{figure}

In this section, we carry out a comprehensive experiment on \texttt{CIFAR-10} in the targeted setting (more difficult attack). In our experimental setup, we use an evaluation set of 9000 different pairs of the source image and target classes (1000 images distributed evenly in 10 different classes against 9 target classes) to compare \SpaSB (500 queries)  with \SpaEvoAtt (2k queries) introduced in~\citep{Vo2022}. We compare ASR of different methods across different sparsity thresholds. The results in Figure \ref{fig:sparesb versus adapted attacks} demonstrate that our attack significantly outperforms SparseEvo. This is expected because SparseEvo is a decision-based attack and has only access to predicted labels.

\textbf{Alternative Loss.~} 
We acknowledge that \citet{Vo2022} may point out an alternative fitness function based on output scores by replacing optimizing 
distortion with optimizing loss. However, they did not evaluate their attack method with an alternative fitness function in score-based setting. Employing this alternative fitness function may not obtain a low sparsity level because minimizing the loss does not surely result in a reduction in the number of pixels. Additionally, the Binary Differential Recombination (BDR) in~\citep{Vo2022} is designed for optimizing $l_0$ distortion not a loss objective (\ie alters perturbed pixels to non-perturbed pixels which is equivalent to minimizing distortion). Hence, naively adapting \SpaEvoAtt \citep{Vo2022} to score-based settings may not work well.

To demonstrate that, we conduct an experiment on \texttt{CIFAR-10} using the same experimental setup (same evaluation set of 9000 image pairs and a query budget of 500) described above. 
\begin{itemize}
    \item First approach, we adapted the attack method in~\citep{Vo2022} to the score-based setting with an alternative fitness function for minimizing loss based on the output scores. We observed this attack always fails to yield an adversarial example with a sparsity level below 50\%.
    \item Second approach, we adapted \SpaEvoAtt by employing the alternative fitness function, synthetic color image and slightly modifying BDR. Our results in Table \ref{table: decision-based attack with alternative loss} show that the adapted \SpaEvoAtt can create sparse adversarial examples but is unable to achieve a comparable performance to \SpaSB.
\end{itemize}

Overall, even with significant improvements, the sparse attack proposed in~\citep{Vo2022} with an alternative fitness function does not achieve as good performance as \SpaSB with a low query budget.

\begin{table}[htp]
\caption{ASR comparison between our proposal and \SpaEvoAtt (Alternative Loss) on \texttt{CIFAR-10}.}
            \begin{center}
		\resizebox{0.60\linewidth}{!}{
		\begin{tabular}{c|cc}
            \toprule
            {Sparsity}& Our Proposal & \SpaEvoAtt (Alternative Loss) \\ 
            \hline \hline
            {1.0$\%$}&\textbf{68.21$\%$} &{54.78$\%$}\\ 
            %\midrule
            {2.0$\%$} &\textbf{90.24$\%$}&{68.75$\%$}\\
            %\midrule
            {2.9$\%$} &\textbf{96.59$\%$}&{74.0$\%$}\\ 
            %\midrule
            {3.9$\%$} &\textbf{98.48$\%$}&{78.56$\%$}\\
            \bottomrule 
            \end{tabular}
            }
            \end{center}
		\label{table: decision-based attack with alternative loss}
  \vspace{-4mm}
\end{table}

\noindent\textbf{Clarifying Differences Between \SpaSB and \SpaEvoAtt (Decision-Based Sparse Attack).~}
\label{apdx:Clarification about The Distinction}
\citet{Vo2022} develops an algorithms for a sparse attack but assumes a decision-based setting. We compared agianst the attack method and provided results in Figure~\ref{fig:sparesb versus adapted} in the main article. Although both works aim to propose sparse attacks, key differences exist, as expected; we explain these differences below:
\begin{itemize}
    \item While both works discuss how they reduce dimensionality (a dimensionality reduction scheme) leading to a reduction in search space from $C \times H \times W$ to $H \times W$, \citet{Vo2022} neither propose a New Problem Formulation nor give proof of showing the equivalent between the original problem in Equation (1) and the New Problem Formulation in Equation (2) as we did in Section \ref{sec:Problem Formulation} and Appendix \ref{apdx:reformulate the original problem}.
    \item Our study and \citet{Vo2022} propose similar terms binary matrix $\vu$ versus binary vector $\vv$ as well as an interpolation between $\vx$ and $\vx'$. However, a binary vector $\vx$ in~\citep{Vo2022} evolves to reduce the number of 1-bits while a binary matrix $\vu$ in our study maintains a number of 1-elements during searching for a solution.
    \item We can find a similar notion of employing a starting image (a pre-selected image from a target class) in~\citep{Vo2022} or synthetic color image (pre-defined by randomly generating) in our study. However, it is worth noting that applying a synthetic color image to \citet{Vo2022} does not work in the targeted setting. For instance, to the best of our knowledge, there is no method can generate a synthetic color image that can be classified as a target class so the method in~\citep{Vo2022} is not able to employ a synthetic color image to inialize a targeted attack. In contrast, employing a starting image as used in~\citep{Vo2022} does not result in query-efficiency as shown in Table \ref{table: brusliattack with starting image}, especially at low sparsity levels.

\end{itemize}
Overall, although the score-based setting is less strict than the decision-based setting, our study is not a simplified version of \citet{Vo2022}.

\begin{table}[h]
\caption{A comparison of ASR between our proposal (Synthetic Color Image) and employing a starting image as in~\citep{Vo2022} on \texttt{CIFAR-10}.}
            \begin{center}
		\resizebox{0.450\linewidth}{!}{
		\begin{tabular}{c|cc}
            \toprule
            {Sparsity}& Our Proposal & Use starting image \\ 
            \hline \hline
            {1.0$\%$}&\textbf{68.21$\%$} &{62.68$\%$}\\ 
            %\midrule
            {2.0$\%$} &\textbf{90.24$\%$}&{87.17$\%$}\\
            %\midrule
            {2.9$\%$} &\textbf{96.59$\%$}&{94.37$\%$}\\ 
            %\midrule
            {3.9$\%$} &\textbf{98.48$\%$}&{97.17$\%$}\\
            \bottomrule 
            \end{tabular}
            }
            \end{center}
		\label{table: brusliattack with starting image}
  \vspace{-4mm}
\end{table}

\subsection{Impact of the Bayesian Framework Based Search (Adapted \SparseRS Using Synthetic Images)}\label{apdx:sparseRSSynthetic}
In this section, we conduct an experiment on \texttt{ImageNet} and in targeted settings to compare the performance of our method and adapted \SparseRS employing synthetic images. Specifically, we replace the update step in \SparseRS by fixing the colors to be changed to the ones in a synthetic image. We employ the same evaluation dataset as discussed in Section \ref{subsec:Decision-Based and l0 Adapted Attacks}. 

The results in Table \ref{table:adapted SparseRS using synthetic images} demonstrate that adapted \SparseRS is less query-efficient than \SpaSB and even the original \SparseRS. In order words, the adapted \SparseRS does not benefit from space reduction by employing synthetic images. A possible reason is that the stochastic pixel selection scheme in \SparseRS does not leverage historical information on pixel manipulation to determine high and low-influential pixels for preservation or replacement. Therefore, \textit{solely employing synthetic images without our proposed learning framework based on historical information regarding pixel manipulation is not found to achieve high query efficiency}. 

\begin{table}[htp]
\caption{ASR at different sparsity levels across different query budgets (higher is better). A comprehensive comparison among different attacks (\SparseRS and \SpaSB) against various Deep Learning models on \texttt{ImageNet} in the targeted setting.}
\label{table:adapted SparseRS using synthetic images}
\begin{center}
\resizebox{0.70\linewidth}{!}{
\begin{tabular}{c|ccc}
\toprule
Query &\SparseRS & \SparseRS(Synthetic Images) &\SpaSB \\ 
\hline \hline
 %====================================================================================
% \midrule
\multicolumn{4}{c}{Sparsity = 0.4$\%$}\\
\cline{1-4}
{4000}& {49.9}$\%$& {49.2}$\%$   & \textbf{57.3}$\%$\\
{6000} & {65.5}$\%$& {63.5}$\%$  & \textbf{69.4}$\%$ \\ 
{8000}& {74.1}$\%$ & {73.6}$\%$ & \textbf{77.3}$\%$ \\
{10000} & {79.1}$\%$ & {79.3}$\%$ & \textbf{82.7}$\%$\\
%====================================================================================
\midrule
\multicolumn{4}{c}{Sparsity = 0.6$\%$}\\
\cline{1-4}
{4000}& {59.6}$\%$& {58.7}$\%$   & \textbf{75.1}$\%$\\
{6000} & {74.0}$\%$& {73.8}$\%$  & \textbf{86.3}$\%$ \\ 
{8000}& {85.0}$\%$& {85.0}$\%$  & \textbf{90.3}$\%$\\
{10000} & {90.9}$\%$& {90.0}$\%$  & \textbf{93.0}$\%$\\
%====================================================================================
\midrule
\multicolumn{4}{c}{Sparsity = 0.8$\%$}\\
\cline{1-4}
{4000}& {65.8}$\%$& {62.9}$\%$  & \textbf{84.3}$\%$\\
{6000} & {79.2}& {78.7}$\%$  & \textbf{90.6}$\%$\\ 
{8000} & {87.9}$\%$& {87.7}$\%$  & \textbf{94.3}$\%$\\
{10000} & {93.4}$\%$& {92.9}$\%$  & \textbf{96.4}$\%$\\
%====================================================================================
\midrule
\multicolumn{4}{c}{Sparsity = 1.0$\%$}\\
\cline{1-4}
{4000}& {69.3}$\%$& {67.3}$\%$  & \textbf{88.6}$\%$\\
{6000} & {82.1}& {81.8}$\%$  & \textbf{94.2}$\%$\\ 
{8000} & {89.8}$\%$& {89.7}$\%$  & \textbf{96.8}$\%$\\
{10000} & {94.3}$\%$& {93.8}$\%$  & \textbf{97.7}$\%$\\
\bottomrule 
\end{tabular}
}
\end{center}
\end{table}

\subsection{$l_0$ Adaptations of Dense Attacks}

\textbf{Adapted $l_0$ Attacks (White-box).~}\hl{To place the blackbox attack results into context by using a whitebox baseline and to provide a baseline for blackbox attack adaptations to $l_0$}, we explore a strong white-box $l_0$ attack. \hl{We used \pgdlO~\citep{Croce2019}---the attack is adaptation of the well-known} PGD~\citep{Madry2017} attack. To this end, we compare \SpaSB with white-box adapted $l_0$ attack \pgdlO using the same evaluation set from \texttt{CIFAR-10} as decision-based attacks. 

The results in Figure \ref{fig:sparesb versus adapted attacks} demonstrate that our attack significantly outperforms \pgdlO at low sparsity threshold and is comparable to \pgdlO at high level of sparsity.  Surprisingly, our method outweighs white-box, adapted $l_0$ attack \pgdlO. It is worth noting that there is no effective projection method to identify the pixels that can satisfy sparse constraint and solving the $l_0$ projection problem also encounters an NP-hard problem. Additionally, the discrete nature of the $l_0$ ball impedes its amenability to continuous optimization \citep{Croce2022}.

\textbf{Adapted $l_0$ Attacks (Decision-based, \hl{Black-box}).~}It is interesting to adapt $l_2$ attacks such as HSJA \citep{Chen2020a}, QEBA \citep{Li2020}, or CMA-ES \citep{Dong2020} method for face recognition tasks to $l_0$ attacks. Consequently, we adopted the HSJA method to an $l_0$ constraint algorithm called $l_0$-HSJA to conduct a study. For $l_0$-HSJA, we follow the experiment settings and adapted $l_0$-HSJA in~\citep{Vo2022} and refer to \citep{Vo2022} for more details. Notably, the same approach could be adopted for QEBA ~\citep{Li2020}.  The results in Table \ref{table:BruSLi vs adapted attacks} below illustrate the average sparsity for 100 randomly selected source images, where each image was used to construct a sparse adversarial sample for the 9 different target classes on \texttt{CIFAR-10}---hence we conducted 900 attacks or used 900 source-image-to-target-class pairs. The average sparsity across different query budgets is higher than 90\% even up to 20K queries. Therefore, the ASR is always 0\% at low levels of sparsity (\ie 4\%) (shown in Figure \ref{fig:sparesb versus adapted attacks}). These results confirm the findings in~\citep{Vo2022} and demonstrate that $l_0$-HSJA (20K queries) is not able to achieve good sparsity (lower is better) when compared with our attack method. Consequently, applying an $l_0$ projection to decision-based dense attacks does not yield a strong sparse attack.

Similar to the problem of \pgdlO, adapted $l_0$-HSJA has to determine a projection that minimizes $l_0$ (the minimum number of pixels) such that the projected instance is still adversarial. To the best of our knowledge, no method in a decision-based setting is able to effectively determine which pixels can be selected to be projected such that the perturbed image does not cross the unknown decision boundary of the DNN model. Solving this projection problem may also lead to another NP-hard problem \citep{Modas2019, Dong2020} and hinders the adoption of these dense attack algorithms to the $l_0$ constraint. Consequently, any adapted method, such as HSJA or other dense attacks, is not capable of providing an efficient method to solve the combinatorial optimization problem faced in sparse settings.

\begin{table}[htp]
\caption{Mean sparsity at different queries for a targeted setting. A sparsity comparison between $l_0$-HSJA on a set of 100 image pairs on \texttt{CIFAR-10}.}
\label{table:BruSLi vs adapted attacks}
\begin{center}
\begin{tabular}{c|c|c|c|c|c}
\toprule
{Queries}&{4000} & {8000} & {12000} & {16000} & {20000} \\
\hline \hline
{$l_0$-HSJA}& {93.66\%} & {94.73\%} & {95.88\%} & {96.74\%} & {96.74\%} \\ 
\bottomrule 
\end{tabular}
\end{center}
\end{table}

\subsection{\hl{Comparing \SpaSB With One-Pixel Attack}}
\label{apdx:Comparing with One-Pixel Attack}
\hl{In this section, we conduct an experiment to compare \SpaSB with the One-Pixel Attack~\citet{Su2019}. We conduct an experiment with 1000 correctly classified images by ResNet18 on CIFAR10 in untargeted settings (notably the easier attack, compared to targeted settings) using ResNet18  These images are evenly distributed across 10 different classes. We compare ASR between our attack and One-Pixel at different budgets e.g. one, three and five perturbed pixels. For One-Pixel attack\footnote{https://github.com/Harry24k/adversarial-attacks-pytorch}, we used the default setting with 1000 queries. To be fair, we set the same query limits for our attack. The results in Table \ref{table:compare with one-pixel} show that our attack outperforms the One-Pixel attack across one, three and five perturbed pixels, even under the easier, untargeted attack setting.}

\begin{table}[htp]
\caption{\hl{ASR comparison (higher $\uparrow$ is stronger) between One-Pixel and \SpaSB against ResNet18 on \texttt{CIFAR-10}.}}
            \begin{center}
		\resizebox{0.50\linewidth}{!}{
		\begin{tabular}{c|cc}
            \toprule
            {\hl{Perturbed Pixels}}& \hl{One-Pixel} & \hl{\SpaSB} \\ 
            \hline \hline
            \hl{1 pixel}& \hl{19.5$\%$} & \hl{\textbf{27.9$\%$}}\\ 
            %\midrule
            \hl{3 pixel} & \hl{41.9$\%$}& \hl{\textbf{69.9$\%$}}\\
            %\midrule
            \hl{5 pixel} & \hl{62.3$\%$}&\hl{\textbf{86.4$\%$}}\\ 
            \bottomrule 
            \end{tabular}
            }
            \end{center}
		\label{table:compare with one-pixel}
  \vspace{-4mm}
\end{table}

\subsection{Bayesian Optimization}

We are interested in the application of Bayesian Optimization for high-dimensional, mix search space. Recently, \citep{Wan2021} has introduced \bo, a Bayesian Optimization for categorical and mixed search spaces, demonstrating that this method is efficient and better than other Bayesian Optimization methods in searching for adversarial examples in score-based settings. Therefore, we study and compare our method with \bo \textit{in the vision domain and the application of seeking sparse adversarial examples}. We note that:
\begin{itemize}
    \item \bo solves problem \ref{eq:problem formulation} directly by searching for altered pixel positions and the colors for these pixels. In the meanwhile, our method aims to address problem \ref{eq:problem formulation with binary vector}, which is reformulated to reduce the dimensionality and complexity of the search space significantly. In general, \bo aims to search for both color values and pixel positions, whilst \SpaSB only seeks pixel locations. 
    \item To handle high dimensional search space in an image task, \bo employs different downsampling/upsampling techniques. It first downscales the image and searches over a low-dimensional space, manipulates and then upscales the crafted examples. Unlike \bo, our method--\SpaSB--does not reduce dimensionality by downsampling the original search space but only seeks pixels in an image (source image) and replaces them with corresponding pixels from a synthetic color image (a fixed and pre-defined image) (see Appendix \ref{apdx:Search space reformulation} for our analysis of dimensionality reduction).
    \item \bo is not designed to learn the impact of pixels on the model decisions but treats all pixels equally, whereas \SpaSB aims to explore the influence of pixels through the historical information of pixel manipulation. 
\end{itemize}
\begin{figure}[htp]
    \begin{center}
        \includegraphics[scale=0.6]{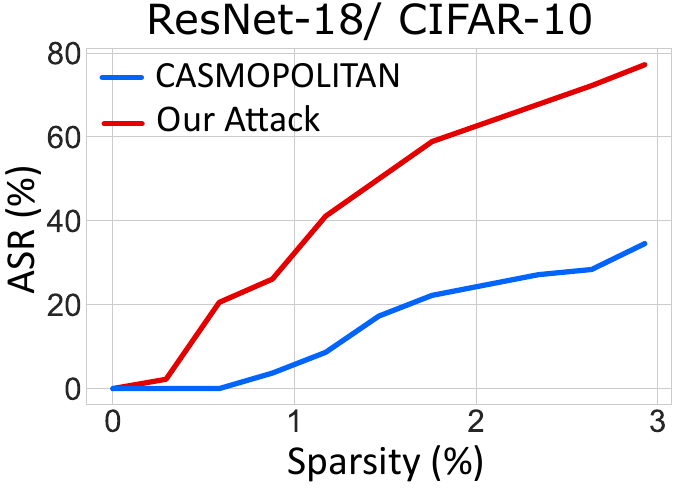}
        \caption{\textbf{Targeted attacks} on \texttt{CIFAR-10} with a query budget of 250. ASR comparisons between \SpaSB and \bo (\textbf{\textit{Bayesian Optimization}}). 
        }
	\label{fig:sparesb versus BO attacks}
    \end{center}
\end{figure}
We use the code\footnote{https://github.com/xingchenwan/Casmopolitan} provided in~\citep{Wan2021} and follow their default settings. We evaluate both \SpaSB and \bo on an evaluation set of 900 pairs of a source image and a target class from \texttt{CIFAR-10} (100 correctly classified images distributed evenly in 10 different classes versus the 9 other classes as target classes for each image) with a query budget of 250. The results in Figure \ref{fig:sparesb versus BO attacks} show that \SpaSB consistently and pragmatically outperforms \bo across different sparsity levels. This is because:
\begin{itemize}
    \item The mixed search space in the vision domain, particularly in sparse adversarial attacks, is still extremely enormous even if downsampling to a lower dimensional search space. It is because \bo still needs to search for a color value for each channel of each pixel from a large range of values (see Appendix \ref{apdx:Search space reformulation} for our analysis of dimensionality reduction).
    \item 
    Searching in a low-dimensional search space and upscaling back to the original search space may not provide an effective way to yield a strong sparse adversarial perturbation. This is because manipulating pixels in a lower dimensional search space may not have the same influence on model decisions as manipulating pixels in the original search space. Additionally, some indirectly altered pixels stemming from upsampling techniques may not greatly impact the model decisions. 
\end{itemize}

%--------------------------------------------------------------------
\subsection{\hl{A Discussion Between \SpaSB (Adversarial Attack) and B3D (Black-box Backdoor Detection)}}
\label{apdx:discussion between ours and B3D}
\hl{\textit{Natural Evolution Strategies (NES).~} A family of black-box optimization methods that learns a search distribution by employing an estimated gradient on its distribution parameters \citet{Wierstra2008, Dong2021}. NES was adopted for score-based dense ($l_2$ and $l_\infty$ norms) attacks in \citet{Ilyas2018} since they mainly adopted a Gaussian distribution for continuous variables. However, solving the problem posed in sparse attacks involving both discrete and continuous variables leads to an NP-hard problem \citet{Modas2019,Dong2020}. Therefore, naively adopting NES for sparse attacks is non-trivial.} 

\hl{The work B3D \citet{Dong2021}, in a defense for a data poisoning attack or backdoor attack, proposed an algorithm to reverse-engineer the potential Trojan trigger used to activate the backdoor injected into a model. Although the method is motivated by NES and operates in a score-based setting involving both continuous and discrete variables, as with a sparse attack problem, they are designed for completely different threat models (backdoor attacks with data poisoning versus adversarial attacks). Therefore it is hard to make a direct comparison. However, more qualitatively, there are a number of key differences between our approach and those relevant elements in \citet{Dong2021}.}
\begin{enumerate}
    \item \hl{Method and Distribution differences: \citet{Dong2021} learns a search distribution determined by its parameters through estimating the gradient on the parameters of this search distribution. In the meantime, our approach is to learn a search distribution through Bayesian learning. While \citet{Dong2021} employed Bernoulli distribution for working with discrete variables, we used Categorical distribution to search discrete variables.}
    \item \hl{Search space (larger vs. smaller): B3D searches for a potential Torjan trigger in an enormous space as it requires to search for pixels’ position and color. Our approach reduces the search space and only searches for pixels (pixels’ position) to be altered so our search space is significantly lower than the search space used in \citet{Dong2021} if the trigger size is the same as the number of perturbed pixels.}
    \item \hl{Perturbation pattern (square shape vs. any set of pixel distribution): \citet{Dong2021} aims to search for a trigger which usually has a size of $1\times 1, 2\times 2 ~\text{or}~ 3\times 3$ so the trigger shape is a small square. In contrast, our attack aims to search for a set of pixels that could be anywhere in an image and the number of pixels could be varied tremendously (determined by desired sparsity). Thus, the combinatorial solutions in a sparse attack problem can be larger than the one in \citet{Dong2021} (even when we equate the trigger size to the number of perturbed pixels).}
    \item \hl{Query efficiency (is a primary objective vs. not an objective): Our approach aims to search for a solution in a query-efficiency manner while it is not clear how efficient the method is to reverse-engineer a trigger.}

\end{enumerate}

\section{Evaluations Against $l_2, l_\infty$ Robust Models From Robustbench and \hl{$l_1$ Robust Models}}
\label{apdx:Evaluate Against SOTA Robust Models}

\begin{table}[htp]
\caption{A robustness comparison (lower $\downarrow$ is stronger) between \SparseRS and \SpaSB against undefended and defended models employing $l_\infty$, $l_2$ robust models on \texttt{CIFAR-10}. The attack robustness is measured by the degraded accuracy of models under attacks at different sparsity levels.}
            \begin{center}
		\resizebox{1.0\linewidth}{!}{
		\begin{tabular}{c|cc|cc|cc}
            \toprule
            \multirow{2}{*}{Sparsity}& \multicolumn{2}{c}{Undefended Model}&\multicolumn{2}{c}{$l_\infty$-Robust Model} & \multicolumn{2}{c}{$l_2$-Robust Model}\\
            \cline{2-7} 
            &\SparseRS &\SpaSB &\SparseRS &\SpaSB & \SparseRS & \SpaSB\\ 
            \hline \hline
            {0.39$\%$}&{26.5$\%$} &\textbf{24.2$\%$} & {65.9}$\%$ & \textbf{65.0}$\%$ & 84.7$\%$ & \textbf{84.2}$\%$\\ 
            %\midrule
            {0.78$\%$} &{7.8$\%$}&\textbf{6.4$\%$}& {48.1}$\%$  & \textbf{46.0}$\%$& {70.6}${\%}$ &\textbf{68.3}${\%}$\\
            %\midrule
            {1.17$\%$} &{2.5$\%$}&\textbf{2.0$\%$}& {38.1}$\%$ & \textbf{35.1}$\%$  & 57.6${\%}$ &\textbf{54.3}${\%}$\\ 
            %\midrule
            {1.56$\%$} &{0.6$\%$}& \textbf{0.6$\%$}& {28.8}$\%$ & \textbf{26.4}$\%$  & {44.4}${\%}$ & \textbf{43.8}${\%}$\\
            \bottomrule 
            \end{tabular}
            }
            \end{center}
		\label{table:ASR against defense methods-cifar10}
  \vspace{-4mm}
\end{table}

\hl{\textbf{\textit{$l_2, l_\infty$ Robust Models.~}}}To supplement our demonstration of sparse attacks (\SpaSB and \SparseRS) against defended models on \texttt{ImageNet} in Section \ref{subsec: Attack against defended models}, we consider evaluations against SoTA robust models from RobustBench\footnote{https://github.com/RobustBench/robustbench} \citep{croce2020} on \texttt{CIFAR-10}. We evaluate the robustness of sparse attacks (\SpaSB and \SparseRS) against the undefended model ResNet-18 and two pre-trained robust models as follows: 
\begin{itemize}
    \item $l_2$ robust model: “Augustin2020Adversarial-34-10-extra”. This model is a top-7 robust model (over 20 robust models) in the leaderboard of robustbench.
    \item $l_\infty$ robust model: “Gowal2021Improving-70-16-ddpm-100m”. This model is a top-5 robust model (over 67 robust models) in the leaderboard of robustbench.
\end{itemize}
We use 1000 samples correctly classified by the pre-trained robust models and evenly distributed across 10 classes on \texttt{CIFAR-10}.
We use a query budget of 500. We compare the accuracy of different models (undefended and defended models) under sparse attacks across a range of Sparsity from 0.39\% to 1.56\%. Notably, defended models are usually evaluated in the untargeted setting to show their robustness. The range of sparsity in the untargeted setting is usually smaller than the range of sparsity used in the targeted setting. Thus, in this experiment, we use a smaller range of sparsity than the one we used in the targeted setting. Our results in Table\ref{table:ASR against defense methods-cifar10} show that \SpaSB outperforms \SparseRS when attacking undefended and defended models. The results on \texttt{CIFAR-10} also confirm our observations on \texttt{ImageNet}. 

\hl{\textbf{\textit{$l_1$ Robust Models.~}}We also evaluate our attack method’s robustness against $\l_1$ robust models. There are two methods AA-I1 \citet{Croce2021} and Fast-EG-1 \citet{jiang2023} for training $l_1$ robust models. Although \citet{Croce2021} and \citet{jiang2023} illustrated their robustness against $l_1$ attacks, Fast-EG-1 is the current state-of-the-art method (as shown in \citet{jiang2023}). Therefore, we chose the $l_1$ robust model trained by the Fast-EG-1 method for our experiment. In this experiment, we use 1000 images correctly classified by $\l_1$ pre-trained model\footnote{https://github.com/IVRL/FastAdvL1} on \texttt{CIFAR-10}. These images are evenly distributed across ten classes. To keep consistency with previous evaluation, we also use a query budget of 500 and compare the accuracy of the robust model under sparse attacks. The results in Table \ref{table:ASR against L1 robust methods-cifar10} show that our attack outperforms \SpaSB across different sparsity levels. Interestingly, $l_1$ robust models are relatively more robust to sparse attacks then other adversarial training regimes in Table~\ref{table:ASR against defense methods-cifar10}, this could be because $l_0$ bounded perturbations are enclosed in the $l_1$-norm ball.}

\begin{table}[htp]
\caption{\hl{A robustness comparison (lower $\downarrow$ is stronger) between \SparseRS and \SpaSB against undefended and defended models employing $l_1$ robust models on \texttt{CIFAR-10}. The attack robustness is measured by the degraded accuracy of models under attacks at different sparsity levels.}}
            \begin{center}
		\resizebox{0.70\linewidth}{!}{
		\begin{tabular}{c|cc|cc}
            \toprule
            \multirow{2}{*}{\hl{Sparsity}}& \multicolumn{2}{c}{\hl{Undefended Model}}&\multicolumn{2}{c}{\hl{$l_1$-Robust Model}} \\
            \cline{2-5} 
            & \hl{\SparseRS} & \hl{\SpaSB} & \hl{\SparseRS} & \hl{\SpaSB} \\ 
            \hline \hline
            \hl{0.39$\%$}& \hl{26.5$\%$} & \hl{\textbf{24.2$\%$}} & \hl{86.6$\%$} & \hl{\textbf{85.8$\%$}} \\ 
            %\midrule
            \hl{0.78$\%$} & \hl{7.8$\%$}& \hl{\textbf{6.4$\%$}}& \hl{75.8$\%$}  & \hl{\textbf{74.8$\%$}}\\
            %\midrule
            \hl{1.17$\%$} & \hl{2.5$\%$}&\hl{\textbf{2.0$\%$}}& \hl{68.5$\%$} & \hl{\textbf{64.8$\%$}}\\ 
            %\midrule
            \hl{1.56$\%$} & \hl{0.6$\%$}& \hl{\textbf{0.6$\%$}}& \hl{59.4$\%$} & \hl{\textbf{55.9$\%$}}\\
            \bottomrule 
            \end{tabular}
            }
            \end{center}
		\label{table:ASR against L1 robust methods-cifar10}
  \vspace{-4mm}
\end{table}

%--------------------------------------------------------------------
\section{Reformulate the Optimization Problem}
\label{apdx:reformulate the original problem}
Solving the problem in Equaion \ref{eq:problem formulation} lead to an extremely large search space because of searching colors—float
numbers in [0, 1]—for perturbing some pixels. To cope with this problem, we i) reduce the search space by synthesizing a color image $\boldsymbol{x'} \in \{0,1\}^{c \times w \times h}$---that is used to define the color for perturbed pixels in the source image (see Appendix \ref{apdx:Search space reformulation}), 
ii) employ a binary matrix $\boldsymbol{u} \in \{0,1\}^{w \times h}$ to determine positions of perturbed pixels in $\boldsymbol{x}$.  
When selecting a pixel, the colors of all three-pixel channels are selected together. Formally, an adversarial instance $\boldsymbol{\Tilde{x}}$ can be constructed as follows: 
\begin{equation}
\begin{gathered}
\label{eq:ae construction} 
{\boldsymbol{\Tilde{x}}} = (1-\boldsymbol{u}) \vx + \boldsymbol{u}\vx'\\
\end{gathered}
\end{equation}
\noindent\textbf{Proof of The Problem Reformulation.~} Given a source image $\boldsymbol{x} \in [0,1]^{c \times w \times h}$ and a synthetic color image $\boldsymbol{x'} \in \{0,1\}^{c \times w \times h}$. From Equation \ref{eq:ae construction}, we have the following: 
\begin{align*} 
{\Tilde{\vx}} &= (1-\boldsymbol{u}) \vx + \boldsymbol{u} \vx'\\ 
{\Tilde{\vx}}- (1-\boldsymbol{u}) \vx &= \boldsymbol{u} \vx'\\ 
\boldsymbol{u}{\Tilde{\vx}} + (1-\boldsymbol{u}){\Tilde{\vx}} - (1-\boldsymbol{u}) \vx &= \boldsymbol{u} \vx'\\ 
(1-\boldsymbol{u})({\Tilde{\vx}} - \vx) &= \boldsymbol{u}(\vx'-{\Tilde{\vx}})\\ 
\end{align*}
We consider two cases for each pixel here:

\begin{enumerate}
    \item If $u_\text{i,j}=0$: then $(1-u_\text{i,j})({\Tilde{x}}_\text{i,~j} - x_\text{i,~j})=0$, thus ${\Tilde{x}}_\text{i,~j} = x_\text{i,~j}$
    \item If $u_\text{i,j}=1$: then $u_\text{i,j} (x'_\text{i,j}-{\Tilde{x}}_\text{i,~j})=0$, thus ${\Tilde{x}}_\text{i,~j} = x'_\text{i,~j}$
\end{enumerate}

Therefore, manipulating binary vector $\boldsymbol{u}$ is equivalent to manipulating $\Tilde{x}$ according to \ref{eq:ae construction}. Hence, optimizing $L(f({\boldsymbol{\Tilde{x}}}),y^*) $ is equivalent to optimizing $L(f((1-\boldsymbol{u}) \vx + \boldsymbol{u}\vx'),y^*) $.
%------------------------------------------------------------------
% Reformulating the Search Space 
%------------------------------------------------------------------
\section{Analysis of Search Space Reformulation and Dimensionality Reduction} \label{apdx:Search space reformulation}
Sparse attacks aim to search for the positions and color values of these perturbed pixels. For a normalized image, the color value of each channel of a pixel---RGB color value---can be a float number in $[0,1]$ so the search space is enormous. The perturbation scheme proposed in~\citep{Croce2022} can be adapted to cope with this problem. This perturbation scheme limits the RGB values to a set $\{0,~1\}$ so a pixel has eight possible color codes $\{000,~001,~010,~011,~100,~101,~110,~111\}$ where each digit of a color code denotes a color value of a channel. This scheme may result in noticeable perturbations but does not alter the semantic content of the input. However, this perturbation scheme still results in a large search space because it grows rapidly with respect to the image size. To obtain a more compact search space, we introduce a simple but effective perturbation scheme. In this scheme, we uniformly sample at random a color image $\boldsymbol{x'} \in \{0,~1\}^{c \times w \times h}$---\textit{synthetic color image}---to define the color of perturbed pixels in the source image $\boldsymbol{x}$. Additionally, we use a binary matrix for selecting some perturbed pixels in $\boldsymbol{x}$ and apply the matrix to $\boldsymbol{x'}$ to extract color for these perturbed pixels as presented in Appendix \ref{apdx:reformulate the original problem}. Because $\boldsymbol{x'}$ is generated once in advance for each attack and has the same size as $\boldsymbol{x}$, the search space is eight times smaller than using the perturbation scheme in~\citep{Croce2022}. Surprisingly, our elegant proposal is shown to be incredibly effective, particularly in high-resolution images such as \texttt{ImageNet}.

\textbf{Synthetic color image.~} Our attack method does not optimize but pre-specify a synthetic color image $\boldsymbol{x'}$ by using our proposed random sampling strategy in our algorithm formulation. This synthetic image is generated once, dubbed a one-time synthetic color image, for each attack. We have chosen to generate it once rather than optimizing it because:
 \begin{itemize}
     \item We aim to reduce the dimensionality of the search space to find and adversarial example. Choosing to optimize the color image would lead to a difficult combinatorial optimization problem.
     \begin{itemize}
         \item Consider what we presented in Section \ref{sec:Problem Formulation}. To solve the combinatorial optimization problem in Equation \ref{eq:problem formulation}, we might search a color value for each channel of each pixel–a float number in [0,1] and this search space is enormous. For instance, if we need to perturb $n$ pixels and the color scale is $2^m$, the search space is equivalent to $C^{c \times n}_{2^{m \times c \times w \times h}}$.
         
        \item To alleviate this problem, we reformulate problem in Equation \ref{eq:problem formulation} and proposed a search over the subspace $\{0,1\}^{{c \times w \times h}}$. However, the size of this search space is still large.
        
        \item To further reduce the search space, we construct a fixed search space–-a pre-defined synthetic color image $\boldsymbol{x'} \in \{0,1\}^{c \times w \times h}$ for each attack. The search space is now reduced to $C^{n}_{w \times h}$. It is generated by uniformly selecting the color value for each channel of each pixel from \{0, 1\} at random (as presented in Appendix \ref{apdx:Search space reformulation} and \ref{apdx:reformulate the original problem}).
    \end{itemize}
    \item In addition, a pre-defined synthetic color image $\boldsymbol{x'}$-–a fixed search space–-benefits our Bayesian algorithm. If keeping optimizing the synthetic color image $\boldsymbol{x'}$, our Bayesian algorithm has to learn and explore a large number of parameters which is equivalent to $C^{c \times n}_{2^{m \times c \times w \times h}}$ and we might not learn useful information fast enough to make the attack progress.
    \item Perhaps, most interestingly, our attack demonstrates that a solution for the combinatorial optimization problem in Equation \ref{eq:problem formulation} can be found in a pre-defined and fixed subspace. 
 \end{itemize}

 \hl{\textbf{Searching for pixels' position and color concurrently.} In general, changing the color of the pixels in searches led to significant increases in query budgets. In our approach, we aim to model the influence of each pixel bearing a specific color, probabilistically, and learn the probability model through the historical information collected from pixel manipulations. So, we chose not to first search for pixels’ position and search for their color after knowing the position of pixels but we aim to do both simultaneously. In other words, the solution found by our method is a set of pixels with their specific colors.}
%--------------------------------------------------------------
% synthetic color image
%--------------------------------------------------------------
\section{Different Schemes for Generating Synthetic Images} \label{apdx:Randomly Initialize Synthetic Images}
In this section, we analyze the impact of different schemes including different random distributions, maximizing dissimilarity and low color search space.

\textbf{Different random distributions.~}
Since the synthetic color images are randomly generated, we can leverage Uniform or Gaussian distribution or our method. Because the input must be within $[0,1]$, we can sample $\vx'$ from $\mathcal{U}[0,1]$ or $\mathcal{N}(\mu,\,\sigma^{2})$ where $\mu=0.5,\sigma=0.17$. For our method, we uniformly sample at random a color image $\boldsymbol{x'} \in \{0,~1\}^{c \times w \times h}$. In order words, each channel of a pixel receives a binary value $0$ or $1$. The results in Table \ref{table:Initialize Synthetic Images} show that generating a synthetic color image from Uniform distribution is better than Gaussian  distribution but it is worse than our simple method. The experiment illustrates that different schemes of generating the synthetic color image at random have different influences on the performance of \SpaSB and our proposal outweighs other common approaches across different sparsity levels. Particularly at low query budgets (\ie up to 300 queries) and low perturbation budgets (\ie sparsity up to 3\%), our proposal outperforms the other two by a large margin. Therefore, the empirical results show our proposed scheme is more effective in obtaining good performance. Most interestingly, as pointed out by HSJA authors \citep{Chen2020a}, the question of how best to select an initialization method or in their case initial target image remains an open-ended question worth investigating.

\begin{table}[htp]
\caption{Target setting. ASR (higher is better) at different sparsity thresholds in the targeted setting. A comprehensive comparison among different strategies of synthetic color image generation to initialize \SpaSB attack against ResNet18 on \texttt{CIFAR-10}.}
\label{table:Initialize Synthetic Images}
\centering
\resizebox{0.55\linewidth}{!}{
\begin{tabular}{c|ccccc}
\toprule
{Methods}&{Q=100} & {Q=200} & {Q=300} & {Q=400} & {Q=500}\\
\midrule
\multicolumn{6}{c}{Sparsity = 1.0$\%$}\\
\cline{1-6}
Uniform & {32.18}$\%$ & {41.68}$\%$& {48.09} $\%$& 52.38$\%$& {55.48$\%$}\\ 
Gaussian & {21.29}$\%$ & {29.87}$\%$& {35.0} $\%$& 38.72$\%$& {41.53$\%$}\\ 
%& 
\textbf{Ours}  & \textbf{42.32$\%$} & \textbf{54.73$\%$}& \textbf{61.49$\%$} & \textbf{65.33$\%$} & \textbf{68.21$\%$} \\
%----------------------------------------------------------------
\midrule
\multicolumn{6}{c}{Sparsity = 2.0$\%$}\\
\cline{1-6}
Uniform & {54.04$\%$} & {69.08$\%$}& {76.48$\%$} & 80.91$\%$& 83.76$\%$\\
Gaussian & {40.02}$\%$ & {55.17}$\%$& {63.2} $\%$& 68.58$\%$& {72.28$\%$}\\ 
\textbf{Ours} & \textbf{66.01$\%$} & \textbf{79.19$\%$}& \textbf{84.84$\%$} & \textbf{88.27$\%$} & \textbf{90.24$\%$} \\
%----------------------------------------------------------------
\midrule
\multicolumn{6}{c}{Sparsity = 2.9$\%$}\\
\cline{1-6}
Uniform & {65.82$\%$} & {80.62$\%$}& {87.84$\%$} & 91.39$\%$& 93.38$\%$\\ 
Gaussian & {52.4}$\%$ & {69.91}$\%$& {78.42} $\%$& 83.24$\%$& {86.39$\%$}\\ 
\textbf{Ours} & \textbf{75.54$\%$} & \textbf{88.22$\%$}& \textbf{92.91$\%$} & \textbf{95.2$\%$} & \textbf{96.59$\%$} \\
%----------------------------------------------------------------
\midrule
\multicolumn{6}{c}{Sparsity = 3.9$\%$}\\
\cline{1-6}
Uniform & {73.04$\%$} & {86.32$\%$}& {92.33$\%$} & 95.02$\%$& 96.34$\%$\\ 
Gaussian & {61.0}$\%$ & {77.26}$\%$& {84.88} $\%$& 89.63$\%$& {91.94$\%$}\\ 
\textbf{Ours} & \textbf{80.44$\%$} & \textbf{91.24$\%$}& \textbf{95.43$\%$} & \textbf{97.4$\%$} & \textbf{98.48$\%$} \\
\bottomrule 
\end{tabular}}
\end{table}

\textbf{Maximizing dissimilarity.~} There may be different ways to implement your suggestion of generating a synthetic color image $\vx’$ that maximize the dissimilarity between the original  image $\vx$ and $\vx’$. But to the best our knowledge, no effective method can generate a random color image $\vx’$ that maximize its dissimilarity with $\vx$. 

Our approach to this suggestion is to find the inverted color values of $\vx$ by creating an inverted image $\vx_\text{invert}$ to explore color values different from $\vx$. We then find the frequency of these color values (in each R, G, B channel) in $\vx_\text{invert}$. Finally, we generate a synthetic color image $\vx'$ such that the more frequent color values (in R, G, B channels) in $\vx_\text{invert}$ will appear more frequently in $\vx'$. By employing the frequency information of color values in $\vx$, we can create a synthetic color image $\vx'$ that is more dissimilar to $\vx$. In practice, our implementation is described as follow:
\begin{itemize}
    \item Yield the inverted image $\vx_\text{invert}$ = 1 - $\vx$. Note that $\vx \in [0,1]^{c\times w\times h}$
    \item Create a histogram of pixel colors (for each R, G, B channel) to have their frequency in $\vx_\text{invert}$. 
    \item Then we randomly generate a synthetic color image based on the frequency of color values that allows us to maximize the dissimilarity. 
\end{itemize}
The results in Table \ref{table: study inverted synthetic image} show that an approach of maximizing the dissimilarity (using frequency information) yields better performance at low sparsity levels as we discussed in Appendix \ref{apdx:With vs Without Leveraging Prior Knowledge}. However, it does not result in better performance at high levels of sparsity if compared with our proposal. 
\begin{table}[htp]
\caption{ASR comparison between using a synthetic color image uniformly generated at random (our proposal) and maximizing dissimilarity on \texttt{CIFAR-10}.}
            \begin{center}
		\resizebox{0.5\linewidth}{!}{
		\begin{tabular}{c|cc}
            \toprule
            {Sparsity}& Our Proposal & Maximizing Dissimilarity \\ 
            \hline \hline
            {1.0$\%$}&{68.21$\%$} &\textbf{70.16$\%$}\\ 
            {2.0$\%$} &{90.24$\%$}&\textbf{90.75$\%$}\\
            {2.9$\%$} &\textbf{96.59$\%$}&{95.78$\%$}\\ 
            {3.9$\%$} &\textbf{98.48$\%$}&{97.85$\%$}\\
            \bottomrule 
            \end{tabular}
            }
            \end{center}
		\label{table: study inverted synthetic image}
  \vspace{-4mm}
\end{table}

\textbf{Low color search space.~}Instead of reducing the space from 8 color codes to a fixed random one, we consider choosing between 2-4 random colors. That would allow us to search not only in the position space of the pixels but also in their color space without increasing search space significantly. The results in Table \ref{table: study 2-4 random colors} show that expanding color space leads to larger search space. Consequently, this approach may require more queries to search for a solution and results in low ASR, particularly with a small query budget.

\begin{table}[htp]
\caption{ASR comparison between using a fixed random color search space (our proposal) and two or four random color search space on \texttt{CIFAR-10}.}
            \begin{center}
		\resizebox{0.65\linewidth}{!}{
		\begin{tabular}{c|ccc}
            \toprule
            {Sparsity}& Our Proposal & Two Random Colors & Four Random Colors\\ 
            \hline \hline
            {1.0$\%$}&\textbf{68.21$\%$} &{60.11$\%$} & {57.9}$\%$\\ 
            {2.0$\%$} &\textbf{90.24$\%$}&{78.12$\%$}& {78.1}$\%$\\
            {2.9$\%$}&\textbf{96.59$\%$}&{85.89$\%$}& {90.67}$\%$\\ 
            {3.9$\%$} &\textbf{98.48$\%$}&{91.23$\%$}& {95.28}$\%$ \\
            \bottomrule 
            \end{tabular}
            }
            \end{center}
		\label{table: study 2-4 random colors}
  \vspace{-4mm}
\end{table}

\section{\SpaSB under Different Random Seeds}
\label{apdx:Study the performance variation}
It is possible that the initial generated by uniformly selecting the color value for each channel of each pixel from \{0, 1\} at random (as presented in Appendix \ref{apdx:Search space reformulation} and Appendix \ref{apdx:reformulate the original problem}) could impact performance. We investigate this using Monte Carlo experiments. To analyze if our attack is sensitive to our proposed initialization scheme. We run our attack 10 times with different random seeds for each source image and target class pair. This also generates 10 different synthetic color images ($\boldsymbol{x'}$) for each source image and target class pair. We chose an evaluation set of 1000 source images (evenly distributed across 10 random classes) and used each one and our attack to flip the label to 9 different target classes. So we conducted ($1000 \times 9$ source-image-to-target-class pairs) $\times$ 10 (ten because we generated 10 different for each pair) attacks (90K attacks) against ResNet18 on \texttt{CIFAR-10}. We report the min, max, average and standard deviation ASR across the entire evaluation set at different sparsity levels. The results in Table \ref{table:different initialization of synthetic color image} show that our method is invariant to the initialization of $\boldsymbol{x'}$. Therefore, our initialization scheme does not affect the final performance of our attacks reported in the paper. Actually, the more complex task of optimizing $\boldsymbol{x'}$ and devising efficient algorithms to explore the high dimensional search space or the generation of better image synthesizing schemes (initialization schemes) to boost the attack performance leaves interesting works in the future.
\begin{table}[htp]
\caption{ASR (Min, Mean, Max and Standard Deviation) of our attack methods across the entire evaluation set at different sparsity levels with a query budget of 500, with 10 different random seeds 
for each attack on \texttt{CIFAR-10}.}
\label{table:different initialization of synthetic color image}
\begin{center}
\resizebox{0.75\linewidth}{!}{
\begin{tabular}{c|c|c|c|c}
\toprule
{Sparsity}&{ASR (Min)} & {ASR (Mean)} & {ASR (Max)} & {Standard Deviation} \\
\hline \hline
{1\%(10 pixels)}& {68.14 \%} & {68.36 \%} & {68.66\%}& {0.35} \\ 
{2\%(20 pixels)} & {90.24\%} & {90.76\%} & {91.38\%}& 0.48 \\
{2.9\%(30 pixels)} & {96.62\%} & {96.71\%} & {96.78\%}& 0.11 \\
{3.9\%(40 pixels)} & {98.17\%} & {98.35\%} & {98.49\%}& 0.13 \\
\bottomrule 
\end{tabular}}
\end{center}
\end{table}

\section{Effectiveness of Dissimilarity Map}
\label{apdx:With vs Without Leveraging Prior Knowledge}
In this section, we aim to investigate the impact of employing the dissimilarity map as our prior knowledge. 

\textbf{On \texttt{CIFAR-10}.~}Similarly, we conduct another experiment on an evaluation set which is composed of 1000 correctly classified images (from \texttt{CIFAR-10}) evenly distributed in 10 classes and 9 target classes per image. However, to reduce the burden of computation when studying hyper-parameters, we use a query budget of 500. The results in Table \ref{table: study with and without bias-cifar} confirm our observation on \texttt{ImageNet}.
\begin{table}[htp]
\caption{ASR comparison between with and without using Dissimilarity Map on \texttt{CIFAR-10}.}
            \begin{center}
		\resizebox{0.6\linewidth}{!}{
		\begin{tabular}{c|ccc}
            \toprule
            {Sparsity}& With Dissimilarity Map & Without Dissimilarity Map \\ 
            \hline \hline
            {1.0$\%$}&\textbf{68.21$\%$} &{67.16$\%$}\\ 
            {2.0$\%$} &\textbf{90.24$\%$}&{89.42$\%$}\\
            {2.9$\%$} &\textbf{96.59$\%$}&{95.96$\%$}\\ 
            {3.9$\%$} &\textbf{98.48$\%$}&{97.92$\%$}\\
            \bottomrule 
            \end{tabular}
            }
            \end{center}
		\label{table: study with and without bias-cifar}
  \vspace{-4mm}
\end{table}

\textbf{On \texttt{ImageNet}.~}We conduct a more comprehensive experiment in terms of query budget to show more interesting results. In this experiment, we use the same evaluation set of 500 samples from \texttt{ImageNet} used in Section \ref{sec:Evaluations} and in the targeted setting. The results in Table \ref{table:prior knowledge study} show that employing prior knowledge of pixel dissimilarity benefits our attack, particularly at a low percentage of sparsity rather. At a high percentage of sparsity, \SpaSB adopting prior knowledge only achieves a comparable performance to \SpaSB without prior knowledge. Notably, at a sparsity of 0.2$\%$, \SpaSB is slightly worse than \SparseRS. Nonetheless, employing prior knowledge of pixel dissimilarity improve the performance of \SpaSB and makes it consistently outweigh \SparseRS. %The visualization of prior knowledge of pixel dissimilarity is illustrated in  \ref{}. 
\begin{table}[htp]
\caption{ASR at different sparsity thresholds and queries (higher is better) for a targeted setting. A comparison between \SparseRS, \SpaSB (without Dissimilarity Map) and \SpaSB (with Dissimilarity Map) on an evaluation set of 500 pairs of an image and a target class on \texttt{ImageNet}}.
\label{table:prior knowledge study}
\begin{center}
\resizebox{0.9\linewidth}{!}{
\begin{tabular}{c|ccccc}
\toprule
{Methods}&{Q=2000} & {Q=4000} & {Q=6000} & {Q=8000} & {Q=10000}\\
%---------------------------------------------------------
\midrule
\multicolumn{6}{c}{Sparsity = 0.2$\%$}\\
\cline{1-6}
{\SparseRS} & {9.4}$\%$ & {20.6}$\%$& {29.6} $\%$& 33.4$\%$& {38.4$\%$}\\
{\SpaSB (without Dissimilarity Map)} & {8.8}$\%$ & {19.6}$\%$& {27.4} $\%$& 34.4$\%$& {38.2$\%$}\\ 
\textbf{\SpaSB} & \textbf{12$\%$} & \textbf{23.6$\%$}& \textbf{31.6$\%$} & \textbf{36.6$\%$} & \textbf{40.4$\%$} \\
%---------------------------------------------------------
\midrule
\multicolumn{6}{c}{Sparsity = 0.4$\%$}\\
\cline{1-6}
{\SparseRS}& {23.6$\%$} & {48.4$\%$}& {63.0$\%$} & 72.6$\%$& 78.8$\%$\\ 
{\SpaSB (without Dissimilarity Map)}& {30.2$\%$} & {53.4$\%$}& {64.4$\%$} & 73.0$\%$& 78.6$\%$\\ 
\textbf{\SpaSB} & \textbf{33.2$\%$} & \textbf{54.2$\%$}& \textbf{66.8$\%$} & \textbf{76$\%$} & \textbf{82.4$\%$} \\
%---------------------------------------------------------
\midrule
\multicolumn{6}{c}{Sparsity = 0.6$\%$}\\
\cline{1-6}
{\SparseRS}& {29.6$\%$} & {57.6$\%$}& {73.2$\%$} & 85.8$\%$& 92.0$\%$\\ 
{\SpaSB (without Dissimilarity Map)}& {43.6$\%$} & {71.6$\%$}& {85.0$\%$} & 91.8$\%$& 94.6$\%$\\ 
\textbf{\SpaSB} & \textbf{45.4$\%$} & \textbf{75.6$\%$}& \textbf{87.4$\%$} & \textbf{91.8$\%$} & \textbf{94.6$\%$} \\
\bottomrule 
\end{tabular}}
\end{center}
\end{table}

\section{Hyper-parameters, Initialization and Computation Resources} \label{apdx:hyper-parameters}
All experiments in this study are performed on two RTX TITAN GPU ($2 \times 24$GB) and four RTX A6000 GPU ($4 \times 48$GB). We summarize all hyper-parameters used for \SpaSB on the evaluation sets from \texttt{CIFAR-10}, \texttt{STL-10} and \texttt{ImageNet} as shown in Table~\ref{table:hyper-parameter} . Notably, only the initial changing rate $\lambda_0$ is adjusted for different resolution datasets \ie \texttt{STL-10} or \texttt{ImageNet} ; thus, our method can be easily adopted for different vision tasks. Additionally, to realize an attack, we randomly synthesize a color image $\boldsymbol{x'}$ for each attack. At initialization step, \SpaSB randomly creates 10 candidate solutions and choose the best.

\begin{table*}[htp]
\caption{Hyper-parameters setting in our experiments}
\label{table:hyper-parameter}
\begin{center}
\resizebox{0.75\linewidth}{!}{
\begin{tabular}{ c||cc|cc|cc }

\hline
\multirow{2}{*}{Parameters} & \multicolumn{2}{c}{\bf CIFAR-10} &\multicolumn{2}{c}{\bf STL-10} & \multicolumn{2}{c}{\bf ImageNet}\\
\cline{2-3} \cline{4-5} \cline{6-7} & Untargeted & Targeted & Untargeted & Targeted & Untargeted & Targeted\\ \hline
\hline
\bf $m_1$ & 0.24  & 0.24 &0.24  & 0.24 & 0.24 & 0.24\\ 
\bf $m_2$ & 0.997  & 0.997 & 0.997  & 0.997 & 0.997 & 0.997\\ $\lambda_0$ & 0.3  & 0.15 & 0.3  & 0.15 & 0.3  & 0.05\\ 
\bf $\boldsymbol{\alpha}^{\text{prior}}$ & $\boldsymbol{1}$ & $\boldsymbol{1}$ & $\boldsymbol{1}$  & $\boldsymbol{1}$ & $\boldsymbol{1}$  & $\boldsymbol{1}$\\ 
\hline
\end{tabular}}
\end{center}
\end{table*}

\section{Hyper-Parameters Study} \label{apdx:Hyper-parameters and scheme Impacts}
In this section, we conduct comprehensive experiments to study the impacts and the choice of hyper-parameters used in our algorithm. The experiments in this section are mainly conducted on \texttt{CIFAR-10}. For $\lambda_0$, we conduct an additional experiment on \texttt{ImageNet}.

\subsection{The Impact of $m_1, m_2$}
In this experiment, we use the same evaluation set on \texttt{CIFAR-10} mentioned above. To investigate the impact of $m_1$, we set $m_2 = 0.997$ and change $m_1 = 0.2,~0.24,~0.28$. Likewise, we set $m_1 = 0.24$ and change $m_2 = 0.993,~0.997,~0.999$ to study $m_2$. The results in Table \ref{table: study hyper-parameters m1-m2} show that \SpaSB achieves the best results with $m_1 = 0.24$ and $m_2 = 0.997$.
\begin{table}[htp]
\caption{ASR of \SpaSB with different values of $m_1, m_2$ on \texttt{CIFAR-10}.}
            \begin{center}
		\resizebox{0.80\linewidth}{!}{
		\begin{tabular}{c|ccc|ccc}
            \toprule
            \multirow{2}{*}{Sparsity}& \multicolumn{3}{c}{Fixed $m_2=0.997$}& \multicolumn{3}{c}{Fixed $m_1=0.24$}\\
            \cline{2-7} 
            &$m_1=0.2$ &$m_1=0.24$ &$m_1=0.28$&$m_2=0.993$ & $m_2=0.997$ & $m_2=0.999$\\ 
            \hline \hline
            {1.0$\%$}&{67.32$\%$} &\textbf{68.21$\%$} & {67.48}$\%$ & {67.34}$\%$ & \textbf{68.21}$\%$ & {67.21}$\%$\\ 
            {2.0$\%$} &{88.67$\%$}&\textbf{90.24$\%$}& {88.94}$\%$&{89.64}$\%$ & \textbf{90.24}$\%$ & {89.12}$\%$\\
            {2.9$\%$} &{95.37$\%$}&\textbf{96.59$\%$}& {95.54}$\%$ &{96.25}$\%$ & \textbf{96.59}$\%$ & {95.82}$\%$\\
            {3.9$\%$} &{97.24$\%$}& \textbf{98.48$\%$}& {97.68}$\%$&{97.59}$\%$ & \textbf{98.48}$\%$ & {96.21}$\%$\\
            \bottomrule 
            \end{tabular}
            }
            \end{center}
		\label{table: study hyper-parameters m1-m2}
  \vspace{-4mm}
\end{table}

\subsection{The Impact of $\lambda_0$}
\textbf{On \texttt{CIFAR-10}.~}Similarly, we conduct another experiment on the same evaluation set which is composed of 1000 correctly classified images (from \texttt{CIFAR-10}) as described above. We use the same query budget of 500. We use $m_1 = 0.24$ and $m_2 = 0.997$ and change $\lambda_0=0.15$ to study the impact of $\lambda_0$. Our results in Table \ref{table: study hyper-parameters lambda} show that \SpaSB achieves the best results with $\lambda_0=0.15$.
\begin{table}[htp]
\caption{ASR of \SpaSB with different values of $\lambda_0$ on \texttt{CIFAR-10}.}
            \begin{center}
		\resizebox{0.450\linewidth}{!}{
		\begin{tabular}{c|ccc}
            \toprule
            {Sparsity}& $\lambda_0=0.1$ &$\lambda_0=0.15$ &$\lambda_0=0.2$\\ 
            \hline \hline
            {1.0$\%$}&{68.05$\%$} &\textbf{68.21$\%$} & {68.12}$\%$\\ 
            %\midrule
            {2.0$\%$} &{89.38$\%$}&\textbf{90.24$\%$}& {88.33}$\%$  \\
            %\midrule
            {2.9$\%$} &{96.15$\%$}&\textbf{96.59$\%$}& {95.56}$\%$ \\ 
            %\midrule
            {3.9$\%$} &{98.16$\%$}& \textbf{98.48$\%$}& {97.08}$\%$ \\
            \bottomrule 
            \end{tabular}
            }
            \end{center}
		\label{table: study hyper-parameters lambda}
  \vspace{-4mm}
\end{table}

\textbf{On \texttt{ImageNet}.~} We use 500 random pairs of an image and a target class from \texttt{ImageNet} in a targeted setting. We carry on a more comprehensive experiment in terms of query budgets. Figure \ref{fig:Hyper-parameters and schemes Impacts} shows that with different initial changing rates $\lambda_0$, \SpaSB obtains the best results when $\lambda_0$ is small such as 0.03 or 0.05. However, at a small sparsity budget, $\lambda_0=0.03$ often achieves lower ASR than $\lambda_0=0.05$ as shown in Table \ref{table:hyper-parameter study} because it requires more queries to make changes and move towards a solution. Consequently, $\lambda_0$ should not be too small. If increasing $\lambda_0$, \SpaSB reaches its highest ASR slower than using small $\lambda_0$. Hence, the initial changing rate has an impact on the overall performance of \SpaSB. 
\begin{table*}[htp]
	\begin{minipage}{0.32\linewidth}
	\centering
        \includegraphics[scale=0.45]{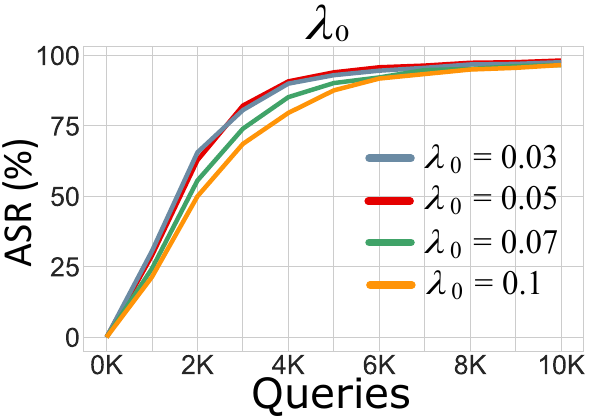}
        \captionof{figure}{ASR versus model queries on $\texttt{ImageNet}$. \SpaSB against ResNet-50 with sparsity of 1.0 $\%$ in a targeted setting to show the impacts of different hyper-parameters on \SpaSB.}
		\label{fig:Hyper-parameters and schemes Impacts}
	\end{minipage} \hfill
	\begin{minipage}{0.65\linewidth}
		\caption{ASR at different sparsity levels and queries (higher is better) in a targeted setting. A comparison between $\lambda_0=0.03$ and $\lambda_0=0.05$ on a set of 500 pairs of an image and a target class on \texttt{ImageNet}}
		\resizebox{1\linewidth}{!}{
\begin{tabular}{c|ccccc}
\toprule
{Initial changing rate}&{Q=2000} & {Q=4000} & {Q=6000} & {Q=8000} & {Q=10000}\\
\midrule
\multicolumn{6}{c}{Sparsity = 0.2$\%$}\\
\cline{1-6}
{$\lambda_0=0.03$} & {10.2}$\%$ & {22.6}$\%$& {29.2} $\%$& 35.6$\%$& \textbf{41.4$\%$}\\ 
{$\lambda_0=0.05$}  & \textbf{12$\%$} & \textbf{23.6$\%$}& \textbf{31.6$\%$} & \textbf{36.6$\%$} & {40.4$\%$} \\
\midrule
\multicolumn{6}{c}{Sparsity = 0.4$\%$}\\
\cline{1-6}
{$\lambda_0=0.03$ }& {31$\%$} & {53.6$\%$}& {65.6$\%$} & 74.2$\%$& 80$\%$\\ 
{$\lambda_0=0.05$} & \textbf{33.2$\%$} & \textbf{54.2$\%$}& \textbf{66.8$\%$} & \textbf{76$\%$} & \textbf{82.4$\%$} \\
\midrule
\multicolumn{6}{c}{Sparsity = 0.6$\%$}\\
\cline{1-6}
{$\lambda_0=0.03$ }& {45.4$\%$} & {75.4$\%$}& {84.6$\%$} & 89.8$\%$& 92.8$\%$\\ 
{$\lambda_0=0.05$} & \textbf{45.4$\%$} & \textbf{75.6$\%$}& \textbf{87.4$\%$} & \textbf{91.8$\%$} & \textbf{94.6$\%$} \\
\bottomrule 
\end{tabular}%}
            }
		\label{table:hyper-parameter study}
	\end{minipage}
\end{table*}

\subsection{The Choice of $\alpha^{prior}$}\label{apdx:the choice of alpha_prior}
\hl{In this section, we discuss the choice of $\boldsymbol{\alpha}^{prior}$ and provide an analysis on the convergence time.}
\begin{itemize}
    \item \hl{$\boldsymbol{\alpha}^{prior}=1$  ($\alpha_i=1$ where $i \in [1,k]$).~ }
In our proposal, we draw multiple pixels (equivalent to multiple elements in a binary matrix u) from the Categorical distribution ($K$ categories) parameterized by $\boldsymbol{\theta} = [\theta_1,\theta_2,...,\theta_K]$. When initializing an attack, we have no prior knowledge of the influence of each pixel that is higher or lower than other pixels on the model’s decision so it is sensible to assume all pixels have a similar influence. Consequently, all pixels should have the same chance to be selected for perturbation (to be manipulated). To this end, the Categorical distribution where multiple pixels are drawn from should be a uniform distribution and $\theta_1=\theta_2=...=\theta_K=\frac{1}{K}$. 

We note that Dirichlet distribution is the conjugate prior distribution of the Categorical distribution. If the Categorical distribution is a uniform distribution, Dirichlet distribution is also a uniform distribution. In probability and statistics, Dirichlet distribution (parameterized by a concentration vector $\boldsymbol{\alpha} =[\alpha_1, \alpha_2 …, \alpha_K]$, each $\alpha_i$ represents the i-th element where $K$ is the total number of elements) is equivalent to a uniform distribution over all of the elements when $\boldsymbol{\alpha} =[\alpha_1, \alpha_2 …, \alpha_K]=[1, 1, …, 1]$. In other words, there is no prior knowledge favoring one element over another. Therefore, we choose $\boldsymbol{\alpha}^{prior}=\boldsymbol{1}$.

    \item \hl{$\boldsymbol{\alpha}^{prior}< 1$  ($\alpha_i<1$ where $i \in [1,k]$).~ }
\hl{We have $\boldsymbol{\alpha^{posterior}}  = \boldsymbol{\alpha}^{prior}  + s^{(t)}$ and $s^{(t)} = (a^{(t)} + z)/(n^{(t)}+z) - 1$. 
So we have $\boldsymbol{\alpha^{posterior}} = \boldsymbol{\alpha}^{prior} + (a^{(t)}+ z)/(n^{(t)}+z) - 1$. Because $(a^{(t)} + z)/(n^{(t)}+z)\leq1$, we cannot choose $\boldsymbol{\alpha}^{prior}< 1$ to ensure that the parameters controlling the Dirichlet distribution are always positive ($\boldsymbol{\alpha^{posterior}}>0$).}

\item \hl{$\boldsymbol{\alpha}^{prior}>1$ ($\alpha_i>1$ where $i \in [1,k]$).~ }
\hl{Since $\boldsymbol{\alpha^{posterior}}$  = $\boldsymbol{\alpha}^{prior}+ (a^{(t)} + z)/(n^{(t)}+z) - 1$ and $0<(a^{(t)} + z)/(n^{(t)}+z)\leq1$, if $\boldsymbol{\alpha}^{prior}\gg1$, in the first few iterations, $\boldsymbol{\alpha^{posterior}}$  almost remains unchanged so the algorithm will not converge. If $\boldsymbol{\alpha}^{prior}>1$, the farther from 1 $\boldsymbol{\alpha}^{prior}$ is, the more subtle the $\boldsymbol{\alpha^{posterior}}$  changes. Now, the update $(a^{(t)} + z)/(n^{(t)}+z)$ needs more iterations (times) to significantly influence $\boldsymbol{\alpha^{posterior}}$. In other words, the proposed method requires more time to learn the Dirichlet distribution (update $\boldsymbol{\alpha^{posterior}}$). Thus, the convergence time will be longer. Consequently, the larger $\alpha_i$ is, the longer the convergence time is.}
\end{itemize}

\section{\SpaSB With Different Schedulers}
\label{apdx:Different Schedulers}
We carry out a comprehensive experiment to examine the impact of different schedulers including cosine annealing and step decay. In this experiment, we use the same evaluation set with 1000 images from \texttt{CIFAR-10} evenly distributed in 10 classes and 9 target classes per image and we use the same query budget (500 queries). The results in Table \ref{table: study different schedulers} show the ASR at different sparsity levels. These results show that our proposed scheduler slightly outperforms all other schedulers. Based on the results, Step Decay or Cosine Annealing schedulers can be a good alternative.

\begin{table}[htp]
\caption{ASR comparison between using a Power Step Decay (our proposal) and other schedulers on \texttt{CIFAR-10}.}
            \begin{center}
		\resizebox{0.550\linewidth}{!}{
		\begin{tabular}{c|ccc}
            \toprule
            {Sparsity}& Our Proposal & Step Decay &Cosine Annealing\\ 
            \hline \hline
            {1.0$\%$}&\textbf{68.21$\%$} &{68.11$\%$} & {68.02}$\%$\\ 
            %\midrule
            {2.0$\%$} &\textbf{90.24$\%$}&{89.34$\%$}& {89.15}$\%$  \\
            %\midrule
            {2.9$\%$} &\textbf{96.59$\%$}&{96.12$\%$}& {95.89}$\%$ \\ 
            %\midrule
            {3.9$\%$} &\textbf{98.48$\%$}&{98.26$\%$}& {98.18}$\%$ \\
            \bottomrule 
            \end{tabular}
            }
            \end{center}
		\label{table: study different schedulers}
  \vspace{-4mm}
\end{table}

\section{Algorithm Pseudocodes}
\label{apdx:Algorithm Pseudocodes}

\vspace{2mm}
\noindent\textbf{Initialization.~}Algorithm \ref{algo:initialization} presents pseudo-code for attack initialization as presented in Section~\ref{sec:Framework for Sparse Attacks}.
\label{apdx:Initialization}
\begin{algorithm}
    \SetKwInOut{KwIn}{Input}
    \DontPrintSemicolon
    \KwIn{source image $\boldsymbol{x}$, synthetic color image $\boldsymbol{x'}$ source label $y$, target label $y_{\text{target}}$\; 
    \quad\quad\quad number of initial samples $N$,perturbation budget $B$, victim model $f_M$ \;
    }
    $\ell \gets \infty$\;
    \For{$i=1$ \KwTo $N$}
    {   
    Generate $\boldsymbol{u}'$ by uniformly enabling $B$ bits of $\mathbf{0}$ at random\;
    $\ell' \leftarrow L(f_M(g(\boldsymbol{u';\vx,\vx'})),y^*)$\;
    \uIf{$\ell' < \ell$}{
        $\boldsymbol{u} \gets \boldsymbol{u'},~\ell \gets \ell'$}
    }
    \KwRet{$\boldsymbol{u},\ell$}
    \caption{Initialization}
    \label{algo:initialization}
\end{algorithm}

\vspace{2mm}
\noindent\textbf{Generation.~}\label{apdx:Generation}Algorithm \ref{algo:generation} presents pseudo-code for generating new data point as presented in Section~\ref{sec:Framework for Sparse Attacks}.
\begin{algorithm}
    \SetKwInOut{KwIn}{Input}
    \DontPrintSemicolon
    \KwIn{probability $\boldsymbol{\theta}$, 
    bias map $\boldsymbol{M}$, mask $\boldsymbol{u}$, changing rate $\lambda$\; 
    }
    $b \leftarrow \lceil (1-\lambda)B \rceil$ \;
    $\vv_1\ldots,\vv_b \sim \operatorname{Cat} (\vv\mid\boldsymbol{\theta}, \boldsymbol{u}=\mathbf{1})$\;
    $\vq_1\ldots,\vv_{B-b} \sim \operatorname{Cat} (\vq\mid\boldsymbol{\theta} \mM, \boldsymbol{u}=\mathbf{0})$\;
    $\boldsymbol{u}^{(t)} = [\vee_{k=1}^b \vv_k^{(t)}] \vee [\vee_{r=1}^{B-b} \vq_k^{(t)}]$\;
    \KwRet{$\boldsymbol{u}$}
    \caption{Generation}
    \label{algo:generation}
\end{algorithm}

\noindent\textbf{Update.~}\label{apdx:update} Algorithm \ref{algo:Update} presents pseudo-code for updating an accepted mask (a solution in round $t$) and estimated $\boldsymbol{\theta}^{(t)}$ as presented in Section~\ref{sec:Framework for Sparse Attacks} and illustrated in Figure \ref{fig:Dirichlet prob density and alpha update}.
\begin{algorithm}
    \SetKwInOut{KwIn}{Input}
    \DontPrintSemicolon
    \KwIn{previous mask and loss$\boldsymbol{u}^{(t-1)},\ell^{(t-1)}, $current mask and loss $\boldsymbol{u}^{(t)},\ell^{(t)}$,small constant $z$, matrices $\boldsymbol{a}^{(t)}, \boldsymbol{n}^{(t)}$\; 
    }
    $\boldsymbol{a} \gets \boldsymbol{a}^{(t)}, \boldsymbol{n} \gets \boldsymbol{n}^{(t)}$\;
    $n_{i,j \in \{[i,j]|(\boldsymbol{u}^{(t-1)} \lor \boldsymbol{u}^{(t)})_{i,j}=1\}}$ increase by 1\;
    \eIf{$\ell^{(t)} < \ell^{(t-1)} $}{
            $\boldsymbol{u} \leftarrow \boldsymbol{u}^{(t)},~\ell \leftarrow \ell^{(t)}$ \;
            }{
            $\boldsymbol{u} \leftarrow \boldsymbol{u}^{(t-1)},~\ell \leftarrow \ell^{(t-1)}$ \; %\tcp*[r]{Update solution}
            $a_{i,j \in \{[i,j]|(\boldsymbol{u}^{(t-1)}  \oplus (\boldsymbol{u}^{(t-1)}  \land \boldsymbol{u}^{(t)}))_{i,j}=1\}}$ increase by 1\;
            }
    $\boldsymbol{s} \leftarrow \frac{\boldsymbol{a}+z}{\boldsymbol{n}+z}$-1\;
    Update $\boldsymbol{\alpha}^{\text{posterior}}$ using $\vs$ and Equation \ref{eq:posterior mean}\;
    Update $\boldsymbol{\theta}$ using $\boldsymbol{\alpha}^{\text{posterior}}$ and Equation \ref{eq:define and update theta}\;
    \KwRet{$\boldsymbol{u}, \ell, \boldsymbol{\theta},\boldsymbol{a},\boldsymbol{n}$}
    \caption{Update}
    \label{algo:Update}
\end{algorithm}

\section{Evaluation Protocol}\label{apdx:Evaluation Protocol}
In this section, we present the evaluation protocol used in this paper. 
\begin{enumerate}
    \item In the targeted attack settings.
    \begin{itemize}
        \item SparseRS \citep{Croce2022} evaluation with ImageNet: Selected 500 source images. But each source image class was flipped to only one random target class using the attack. So that is a total of 500 source-image-to-target class attacks. This evaluation protocol may select the same target class to attack in the 500 attacks conducted. Thus, this could lead to potential biases in the results because some classes may be easier to attack than others.
        
        \item To avoid the problem, in the targeted attack setting, we followed the evaluation protocol used in~\citep{Vo2022}. Essentially, we flip the label of the source image to several targeted classes, this can help address potential biases caused by relatively easier classes getting selected multiple times for a target class.
        
        \item Our evaluation with \texttt{ImageNet}: We randomly selected 200 correctly classified source images evenly distributed among 200 random classes. But, in contrast to SparseRS, we selected 5 random target classes to attack for each source image. In total we did $200 \times 5 = 1000$ source-image-to-target class attacks on \texttt{ImageNet} for targeted attacks.
    \end{itemize}
    \item In the untargeted attack setting (attacks against defended models), we conducted 500 attacks (similar to SparseRS). We randomly selected 500 correctly classified test images from 500 different classes for attacks.
    \item Further, our unique and exhaustive testing with \texttt{CIFAR-10} and \texttt{STL-10} corroborates \texttt{ImageNet} results given the significant amount of resources it takes to attack the high-resolution \texttt{ImageNet} ($224 \times 224$) models.
    \begin{itemize}
        \item For \texttt{STL-10} we conducted 60,093 attacks against each deep learning model (6,677 of all 10,000 images in the test set which are correctly classified versus 9 other classes as target classes for each source image). We used every single test set image in \texttt{STL-10} ($96 \times 96$) in our attacks to mount the exhaustive evaluation where no image from the test set was left out.
        \item  For \texttt{CIFAR-10} ($32 \times 32$) we conducted 9,000 attacks against each deep learning model (1000 random images correctly classified versus the 9 other classes as target classes for each source image).
    \end{itemize}   
    \item For evaluations against a real-world system (GCV) in the significantly more difficult targeted setting (not the untargeted setting), we provide new benchmarks for attack demonstration because we provide a comparison between \SpaSB and the previous attack, \SparseRS. To make it clear, we provide a brief comparison as follows:
    \begin{itemize}
        \item Other related past studies (dense attacks)\citep{Ilyas2018, Guo2019}, showcase an attack against a real-world system but uses 10 attacks. While \citet{Ilyas2018} illustrated only one successful example when carrying out an attack against Google Could Vision. 
        \item Importantly, we did not simply use our method only, as in~\citep{Ilyas2018, Guo2019}  but demonstrated a comparison between \SpaSB and \SparseRS. In practice, we used 10 samples for each attack, so there are 20 attacks.
    \end{itemize}
\end{enumerate}

In general, our evaluation protocol is much stronger than the one used in previous studies. We evaluate on three different datasets \texttt{CIFAR-10}, \texttt{STL-10} (not evaluated in prior attacks) and \texttt{ImageNet} with ResNet-50, ResNet-50 (SIN), Visitation Transformer (not evaluated in prior attacks).

\section{Visualizations of Attack Against Google Cloud Vision} \label{apdx:Visualizations of Attack Against Google Cloud Vision}
\begin{figure}[htp]
  \centering
        \includegraphics[scale=0.63]{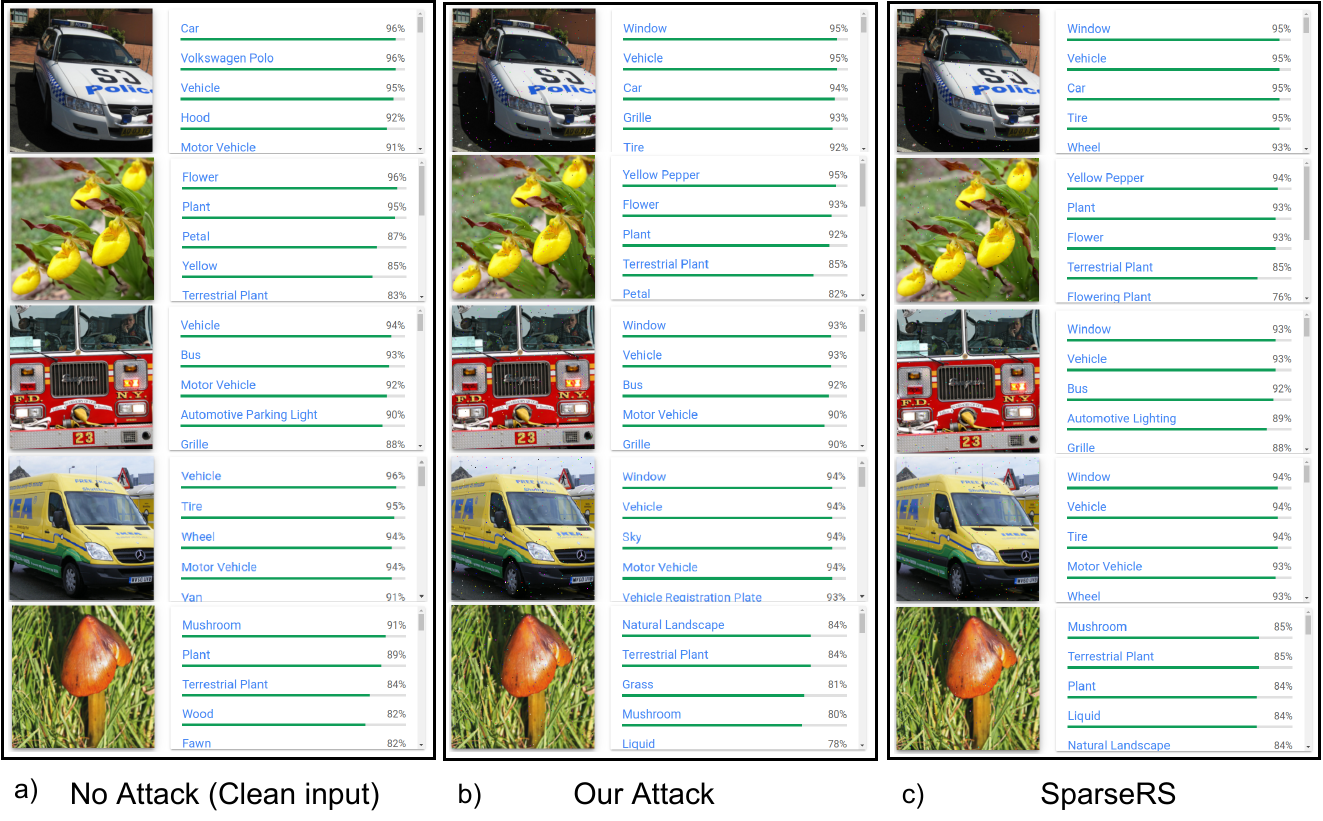}
        \captionof{figure}{a) demonstrates results for clean image (no attack) predicted by Google Cloud Vision (GCV). b) shows the predictions from GCV for adversarial examples crafted successfully by \SpaSB with less than 3,000 queries and sparsity of 0.05 $\%$. c) shows the results from GCV for adversarial examples crafted by \SparseRS with the same sparsity. But \SparseRS needs more queries than \SpaSB to successfully yield adversarial images or fail to attack with query budget up to 5,000 as shown in Table \ref{table:GCV attack results apdx}.
        }
		\label{fig:GCV online demonstration apdx}
\end{figure}

\begin{figure}[htp]
  \centering
        \includegraphics[scale=0.63]{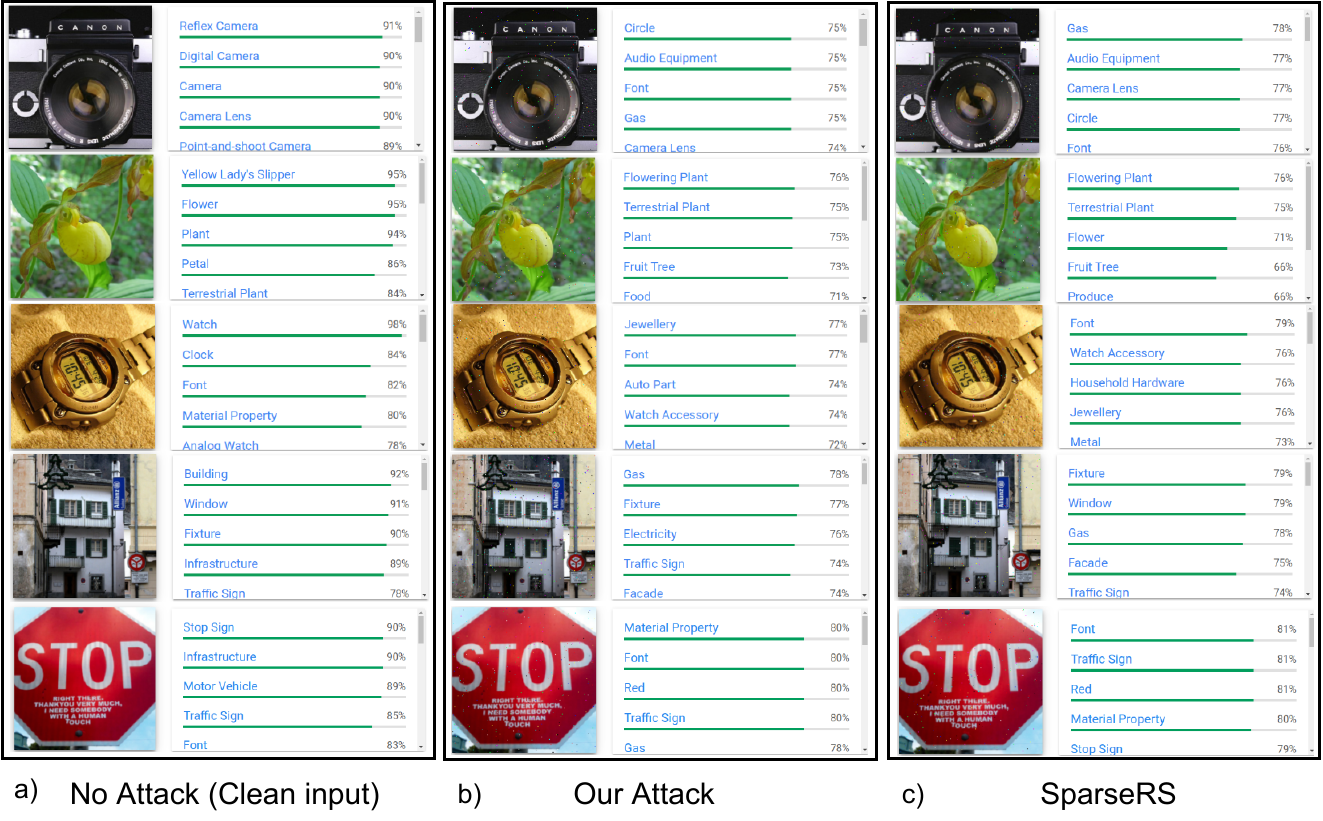}
        \captionof{figure}{a) demonstrates results for clean image (no attack) predicted by Google Cloud Vision (GCV). b) shows the predictions from GCV for adversarial examples crafted successfully by \SpaSB with less than 3,000 queries and sparsity of 0.05 $\%$. c) shows the results from GCV for adversarial examples crafted by \SparseRS with the same sparsity. But \SparseRS need more queries than \SpaSB to successfully yield adversarial images or fail to attack with query budget up to 5,000 as shown in Table \ref{table:GCV attack results apdx}.
        }
		\label{fig:GCV online demonstration1}
\end{figure}

Table \ref{table:GCV attack results apdx} and Figure \ref{fig:GCV online demonstration apdx}, Table \ref{table:GCV attack results apdx1} and Figure \ref{fig:GCV online demonstration1} show our attack against real-world system Google Cloud Vision API.

\begin{table}[htp]
		\centering
		\caption{Demonstration of sparse attacks against GCV in targeted settings. \SpaSB is able to successfully yield adversarial instances for all five examples with less queries than \SparseRS. Especially, for the example of \texttt{Mushroom}, \SparseRS fails to attack GCV within a budget of 5000 queries. Demonstration on GCV API (online platform) is shown in Figure \ref{fig:GCV online demonstration apdx}.}
		\resizebox{1\linewidth}{!}{
		    \begin{tabular}{ cccccc }
            \toprule
            
            {Examples} 
            & \begin{minipage}{.15\textwidth}
                  \includegraphics[scale=0.25]{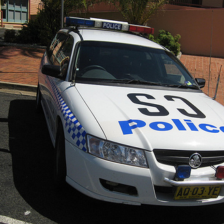}
              \end{minipage}
            & \begin{minipage}{.15\textwidth}
                  \includegraphics[scale=0.25]{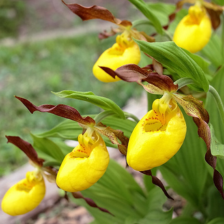}
              \end{minipage}
            & \begin{minipage}{.15\textwidth}
                  \includegraphics[scale=0.25]{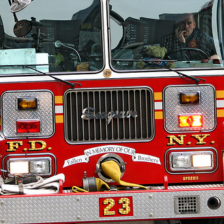}
              \end{minipage}
            & \begin{minipage}{.15\textwidth}
                  \includegraphics[scale=0.25]{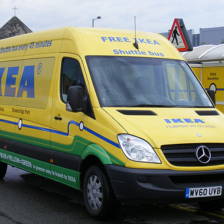}
              \end{minipage}
            & \begin{minipage}{.15\textwidth}
                  \includegraphics[scale=0.25]{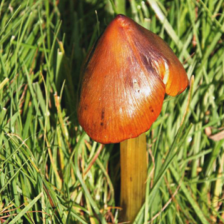}
              \end{minipage}
              \\
            \midrule
            No Attack & \texttt{Car}  & \texttt{Flower}  & \texttt{Fire Truck} & \texttt{Vehicle}  & \texttt{Mushroom} \\ 
            \midrule%\hline
            \multirow{2}{*}{\SpaSB} & \texttt{Window} & \texttt{Yellow Pepper}  & \texttt{Window}& \texttt{Window} & \texttt{Landscape} \\ 
            & ($\boldsymbol{1.8}$K Queries) & ($\boldsymbol{99}$ Queries)  & ($\boldsymbol{328}$ Queries) & ($\boldsymbol{1.83}$K Queries) & ($\boldsymbol{490}$ Queries)
            \\
            \midrule%\hline
            \multirow{2}{*}{\SparseRS} & \texttt{Window}  & \texttt{Yellow Pepper} & \texttt{Window}& \texttt{Window}  & \texttt{Mushroom} \\ 
            & (4.66K Queries) & (211 Queries) & (395 Queries) & (3.3K Queries) & ($>$5K Queries)
            \\
            \bottomrule 
            \end{tabular}
            }
		\label{table:GCV attack results apdx}
\end{table}

\newpage

\begin{table}[htp]
		\centering
		\caption{Demonstration of sparse attacks against GCV in targeted settings. \SpaSB is able to successfully yield adversarial instances for all five examples with less queries than \SparseRS. Especially, for the example of \texttt{Mushroom}, \SparseRS fails to attack GCV within a budget of 5000 queries. Demonstration on GCV API (online platform) is shown in Figure \ref{fig:GCV online demonstration1}.}
		\resizebox{1\linewidth}{!}{
		    \begin{tabular}{ cccccc }
            \toprule
            
            {Examples} 
            & \begin{minipage}{.15\textwidth}
                  \includegraphics[scale=0.25]{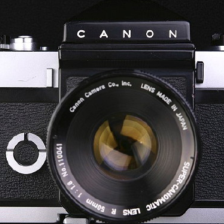}
              \end{minipage}
            & \begin{minipage}{.15\textwidth}
                  \includegraphics[scale=0.25]{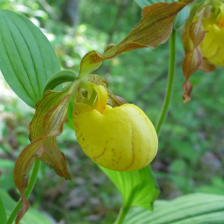}
              \end{minipage}
            & \begin{minipage}{.15\textwidth}
                  \includegraphics[scale=0.25]{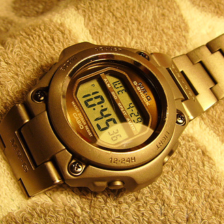}
              \end{minipage}
            & \begin{minipage}{.15\textwidth}
                  \includegraphics[scale=0.25]{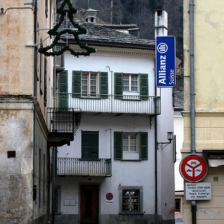}
              \end{minipage}
            & \begin{minipage}{.15\textwidth}
                  \includegraphics[scale=0.25]{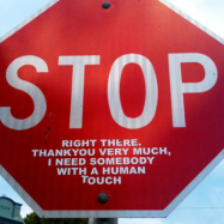}
              \end{minipage}
              \\
            \midrule
            No Attack & \texttt{Reflex Camera}  & \texttt{Y.L.Slipper}  & \texttt{Watch} & \texttt{Building}  & \texttt{Stop Sign} \\ 
            \midrule%\hline
            \multirow{2}{*}{\SpaSB} & \texttt{Circle} & \texttt{Flowering Plant}  & \texttt{Jewellry}& \texttt{Gas} & \texttt{Material P} \\ 
            & ($\boldsymbol{3.8}$K Queries) & ($\boldsymbol{899}$ Queries)  & ($\boldsymbol{2.9}$K Queries) & ($\boldsymbol{983}$ Queries) & ($\boldsymbol{2.77}$K Queries)
            \\
            \midrule%\hline
            \multirow{2}{*}{\SparseRS} & \texttt{Gas}  & \texttt{Flowring Plant} & \texttt{Font}& \texttt{Fixture}  & \texttt{Font} \\ 
            & ($>$5K Queries) & (988 Queries) & ($>$5K Queries) & ($>$5K Queries) & ($>$5K Queries)
            \\
            \bottomrule 
            \end{tabular}
            }
		\label{table:GCV attack results apdx1}
\end{table}

%%%%%%%%%%%%%%%%%%%%%%%%%%%%%%%%%%%%%%%%%%%%%%%%%%%%%%%%%%%%%%%%%%%%%%%%%%%%%%%
\newpage
\section{Visualizations of Sparse Adversarial Examples and Dissimilarity Maps} \label{apdx:Visualization of Sparse Adversarial Examples} 
\hl{In this section, we illustrate:}
\begin{itemize}

    \item \hl{Sparse adversarial examples, sparse perturbation crafted by our methods versus salient region produced by GradCAM method \citet{Selvaraju2017} or attention map produced by a ViT model \citet{Dosovitskiy2021}.}
    
    \item \hl{Sparse adversarial examples crafted by \SpaSB when attacking ResNet-50, ResNet-50 (SIN) and Vistion Transformer.}
    
    \item \hl{Dissimilarity Map produced from a pair of a source and a synthetic color images.}
    
\end{itemize}

\begin{figure*}[htp]
    \begin{center}
        \includegraphics[scale=0.35]{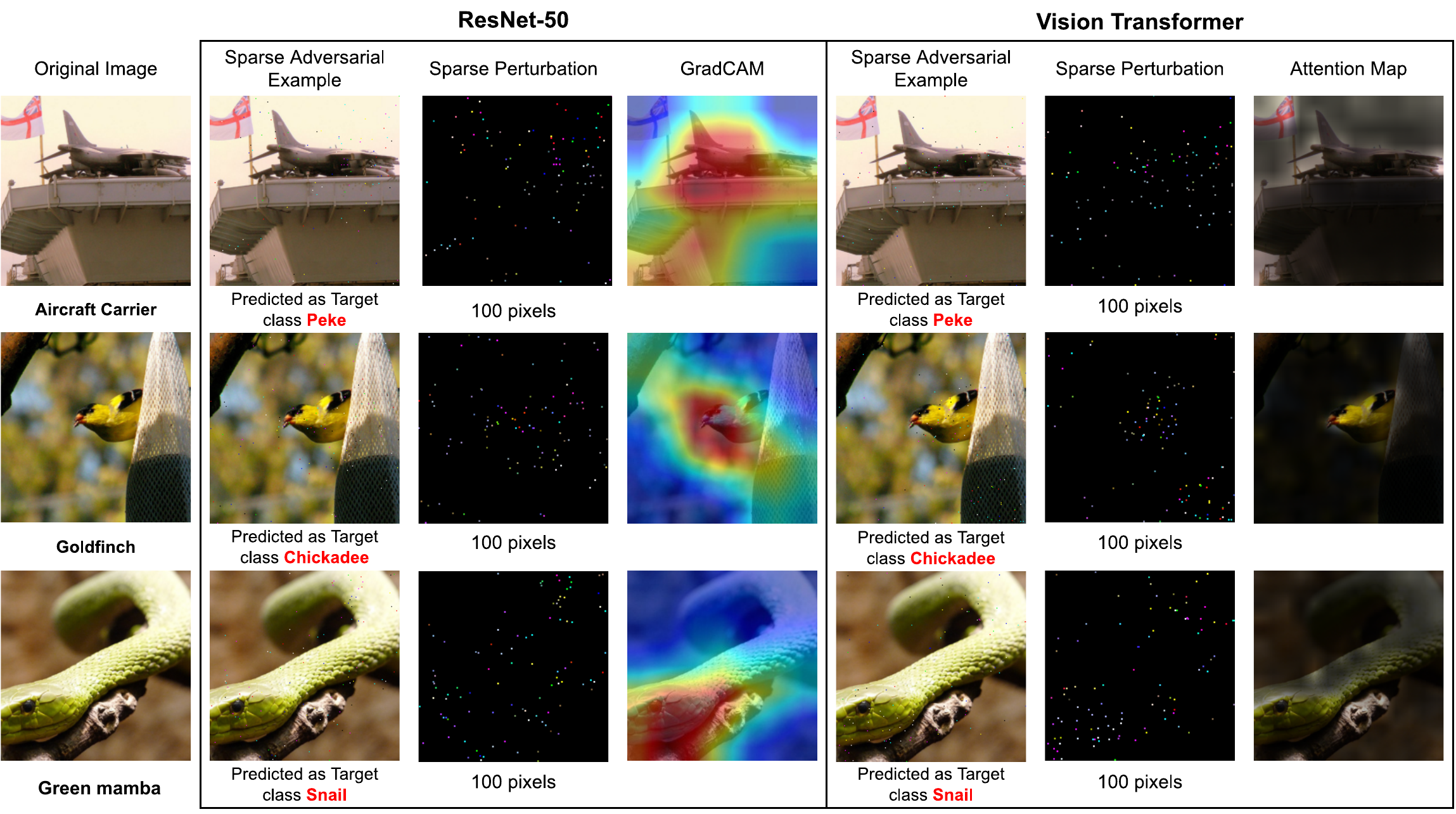}
        \caption{\hl{\textbf{Targeted Attack.~}Visualization of Adversarial examples crafted by \SpaSB with a budget of 10K queries. In the image of sparse perturbation, \textbf{each pixel is zoomed in 9 times ($9\times$)} to make them more visible.}}
        \label{fig:spare adv targeted}
    \end{center}
\end{figure*}

\begin{figure*}[htp]
    \begin{center}
        \includegraphics[scale=0.35]{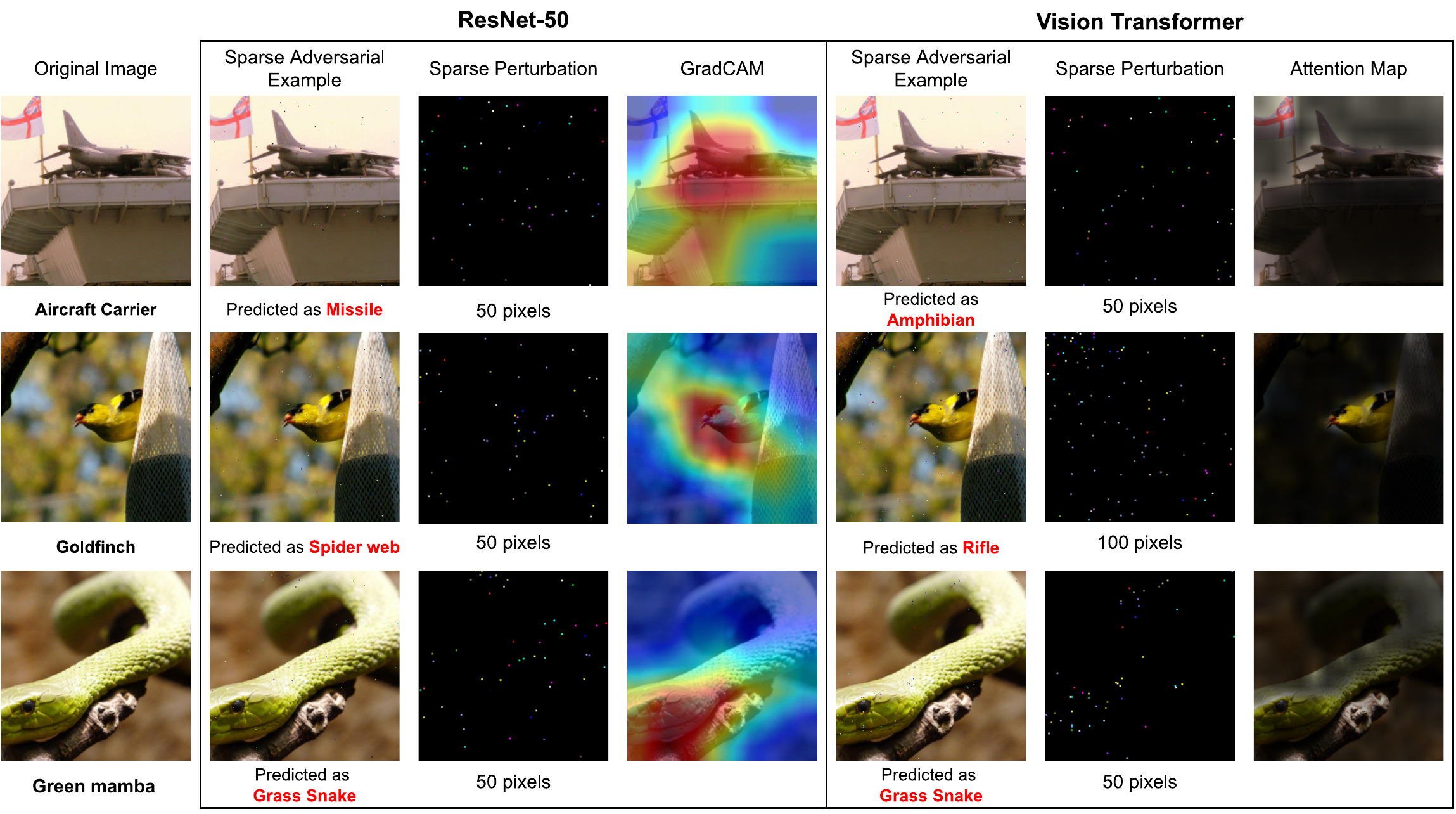}
        \caption{\hl{\textbf{Untargeted Attack.~}Visualization of Adversarial examples crafted by \SpaSB with a budget of 5K queries. In the image of sparse perturbation, \textbf{each pixel is zoomed in 9 times ($9\times$)} to make them more visible.}}
        \label{fig:spare adv untargeted}
    \end{center}
\end{figure*}

\hl{Figure \ref{fig:spare adv targeted} and \ref{fig:spare adv untargeted} illustrate sparse adversarial examples and spare perturbation of images from \texttt{ImageNet} in targeted and untargeted settings. In targeted settings, we use a query budget of 10K, while in untargeted settings, we set a query limit of 5K. We use GradCAM and Attention Map from ViT to demonstrate salient and attention regions. The sparse perturbation $\delta$ is the absolute difference between source images and their sparse adversarial. Formally, sparse perturbations can be defined as $\delta=| \boldsymbol{x}-\boldsymbol{\Tilde{x}}|$.}

\hl{The results show that for ResNet-50, the solutions found do not need to perturb salient regions on an image to mislead the models (both targeted and untargeted attacks). Attacks with ViT models in untargeted settings also lead to a similar observation. Interestingly, for some images e.g. a snake or a goldfinch in Figure~\ref{fig:spare adv targeted}, we observe that a set of perturbed pixels yielded by our method is more concentrated in the attention region of ViT. This seems to indicate some adversarial solutions achieves their objective by degrading the performance of a ViT. This is perhaps not an unexpected observation, given the importance of attention mechanisms to transformer models.}

\begin{figure*}[htp]
    \begin{center}
        \includegraphics[scale=0.9]{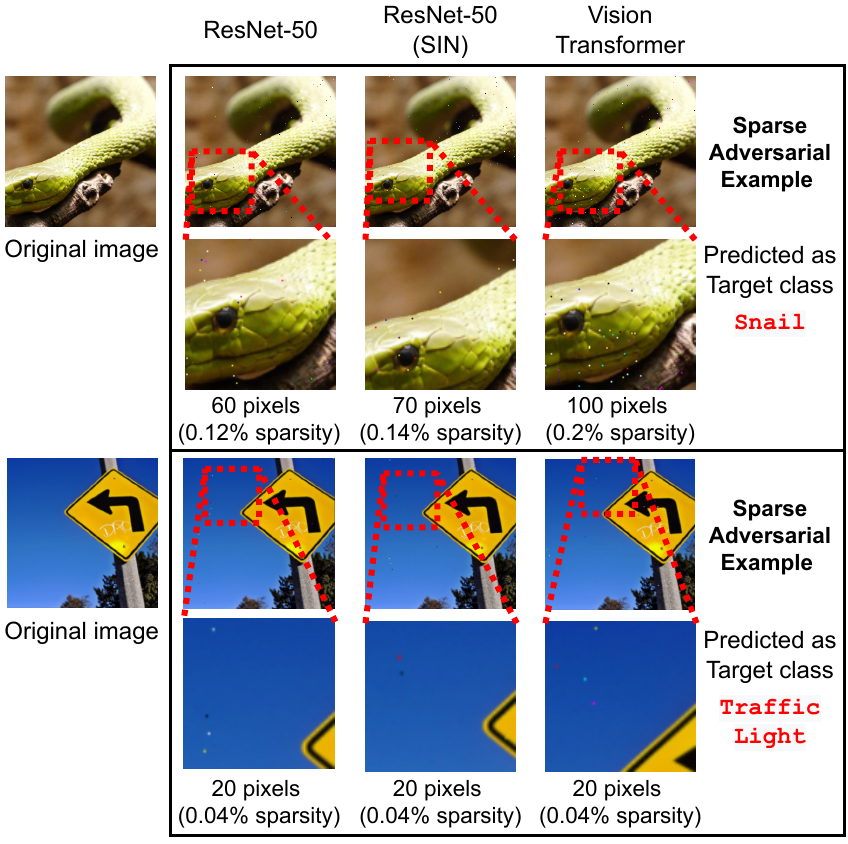}
        \caption{Visualization of Adversarial examples crafted by \SpaSB with a budget of 10K queries.}
        \label{fig:adv-ex visualization}
    \end{center}
\end{figure*}

\begin{figure*}[htp]
    \begin{center}
        \includegraphics[scale=0.9]{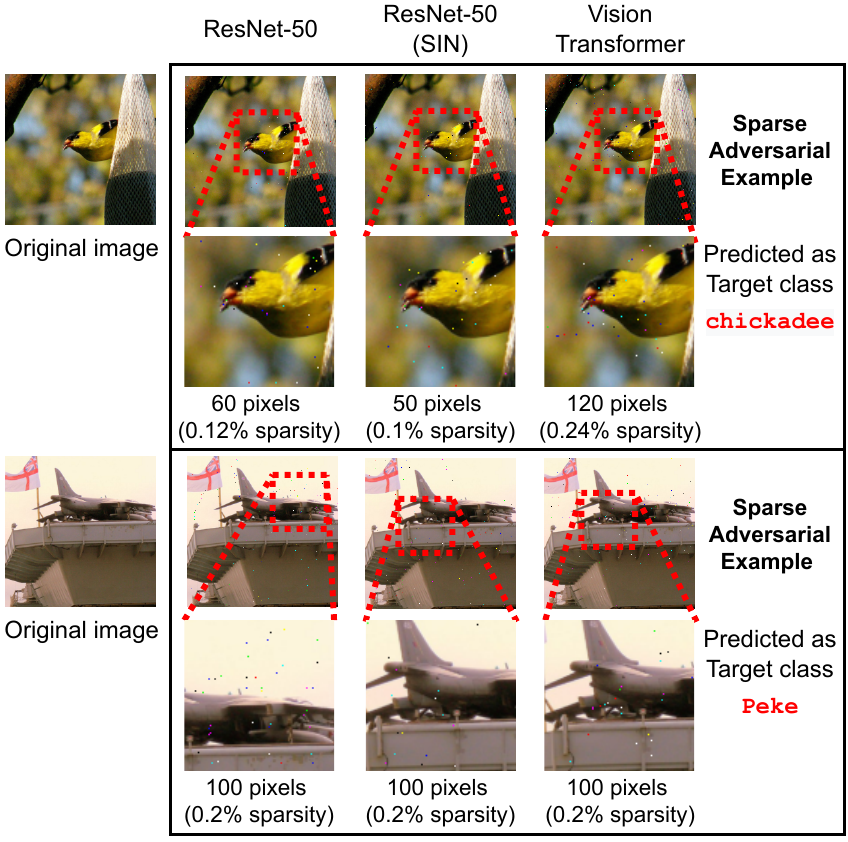}
        \caption{Visualization of Adversarial examples crafted by \SpaSB with a budget of 10K queries.}
        \label{fig:adv-ex visualization1}
    \end{center}
\end{figure*}

\newpage
Figure \ref{fig:adv-ex visualization} and \ref{fig:adv-ex visualization1} demonstrate some examples of adversarial examples yielded by \SpaSB \hl{when attacking different deep learning models (ResNet-50, ResNet-50 (SIN) and Vision Transformer) in targeted settings produced using a 10K query budget.}

\begin{figure}[h!]
    \begin{center}
        \begin{subfigure}
            \centering
            \includegraphics[width=0.9\textwidth]{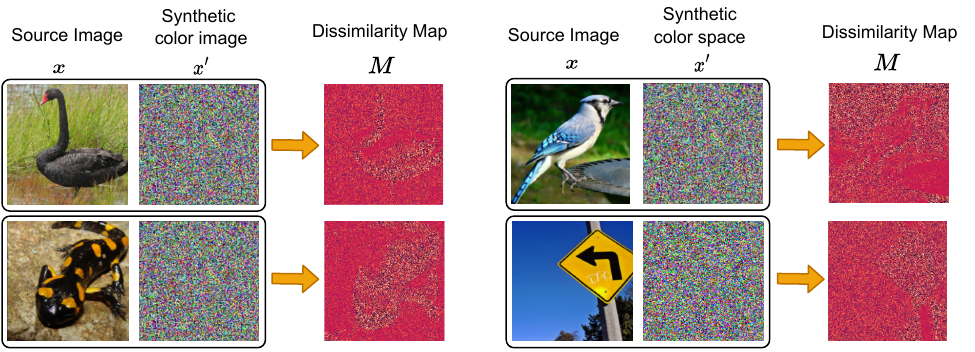}
        
        \end{subfigure}
        
        \begin{subfigure}
            \centering
            \includegraphics[width=0.9\textwidth]{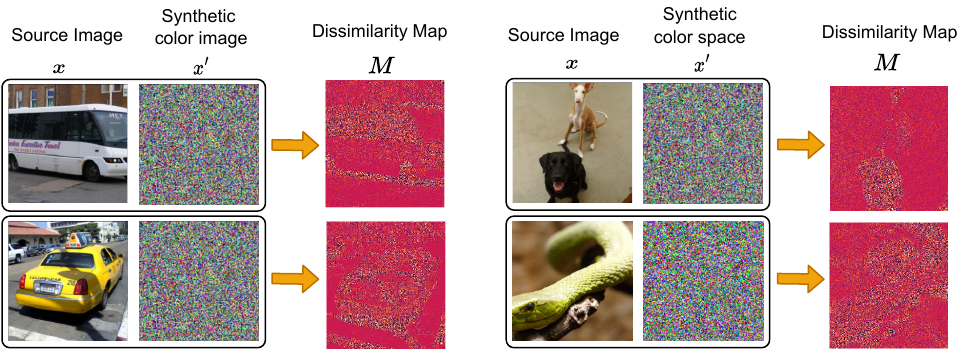}
        
        \end{subfigure}
        \caption{Visualization of Dissimilarity Maps between a source image and a synthetic color image.}
        \label{fig:dissimilarity map}
    \end{center}
\end{figure}

\newpage
Figure \ref{fig:dissimilarity map} illustrates some examples of Dissimilarity Map yielded by a source image and a synthetic color image. 

\end{document}

%% file: math_commands.tex
\usepackage{amsmath,amsfonts,bm}

\def\eqref#1{equation~\ref{#1}}
\def\1{\bm{1}}

\def\vtheta{{\bm{\theta}}}

\def\vn{{\bm{n}}}

\def\vq{{\bm{q}}}

\def\vs{{\bm{s}}}

\def\vu{{\bm{u}}}
\def\vv{{\bm{v}}}

\def\vx{{\bm{x}}}

\def\mM{{\bm{M}}}

\DeclareMathAlphabet{\mathsfit}{\encodingdefault}{\sfdefault}{m}{sl}
\SetMathAlphabet{\mathsfit}{bold}{\encodingdefault}{\sfdefault}{bx}{n}

\newcommand{\E}{\mathbb{E}}

\DeclareMathOperator*{\argmax}{arg\,max}
\DeclareMathOperator*{\argmin}{arg\,min}